\documentclass{article}

\PassOptionsToPackage{numbers, compress}{natbib}
\usepackage[preprint]{neurips_2020}

\usepackage{url}            % simple URL typesetting
\usepackage[utf8]{inputenc} % allow utf-8 input
\usepackage[T1]{fontenc}    % use 8-bit T1 fonts
\usepackage[hidelinks]{hyperref}       % hyperlinks
\usepackage{booktabs}       % professional-quality tables
\usepackage{amsfonts}       % blackboard math symbols
\usepackage{microtype}      % microtypography
\usepackage{amsmath, }
\usepackage{multirow}
\usepackage{graphicx}
\usepackage{xcolor}
\usepackage{epstopdf}
\usepackage{caption}
\usepackage{subcaption}
\usepackage{bm}
\usepackage{pgfplots}
\usepackage{footnote}
\usepackage{stfloats}
\usepackage{placeins}
\usepackage{color,soul}
\usepackage{dsfont}
\usepackage{amssymb}
\usepackage{pbox}
\usepackage{enumerate}
\usepackage{etextools}
\usepackage{mathtools}
\usepackage{amsthm}
\usepackage{tabulary}
\usepackage{dsfont}
\usepackage{pgfplots}
\usepackage{bbm}
\usepackage{comment}
\usepackage{algorithm}
\usepackage[noend]{algorithmic}
\usepackage{subfiles}
\usepackage{wrapfig}

\newtheorem{prop}{Proposition}
\def\Y{{\mathbf Y}}

\def\U{{\mathbf U}}

\def\A{{\mathbf A}}

\def\R{{\mathbb R}}
\def\PP{{\mathbb P}}
\def\RR{{\mathbb R}}

\def\V{{\mathbf V}}

\def\independenT#1#2{\mathrel{\rlap{$#1#2$}\mkern2mu{#1#2}}}
\newcommand\independent{\protect\mathpalette{\protect\independenT}{\perp}}
\def\eqd{\,{\buildrel d \over =}\,} 

\def\Atilde{\tilde{A}}
\def\bAtilde{\widetilde {\mathbf{A}}}
\def\bA{\mathbf{A}}
\def\bY{\mathbf{Y}}
\def\bYhat{\widehat{\mathbf{Y}}}
\def\cov{\text{cov}}

\mathtoolsset{showonlyrefs}

\title{Achieving Equalized Odds by Resampling Sensitive Attributes}

\author{%
  Yaniv Romano \\
  Department of Statistics\\
  Stanford University\\
  \And
  Stephen Bates \\
  Department of Statistics\\
  Stanford University\\
  \And
  Emmanuel J. Cand\`es \\
  Departments of Mathematics \\ 
  and of Statistics\\
  Stanford University\\
}

\begin{document}

\maketitle

\begin{abstract}

We present a flexible framework for learning predictive models that approximately satisfy the equalized odds notion of fairness. This is achieved by introducing a general discrepancy functional that rigorously quantifies violations of this criterion. This differentiable functional is used as a penalty driving the model parameters towards equalized odds. To rigorously evaluate fitted models, we develop a formal hypothesis test to detect whether a prediction rule violates this property, the first such test in the literature. Both the model fitting and hypothesis testing leverage a resampled version of the sensitive attribute obeying equalized odds, by construction. We demonstrate the applicability and validity of the proposed framework both in regression and multi-class classification problems, reporting improved performance over state-of-the-art methods. Lastly, we show how to incorporate techniques for equitable uncertainty quantification---unbiased for each group under study---to communicate the results of the data analysis in exact terms.

%To communicate effectively the data analysis results with decision makers, we build on the recently proposed equalized coverage framework and propose methodologies to construct calibrated and unbiased measures for prediction uncertainty. The constructed prediction intervals/sets are supported by distribution-free finite sample coverage guarantees, and therefore reliably summarize the information extracted from any complex model that is fitted with our learning framework.
\end{abstract}

% \ejc{Moving outstanding issues up front.} 

% \ejc{Do we want to present crime data? This is an extremely sensitive topic at the moment.}\yr{It is the most popular (if not the only) data set that most people use. I will mention that.} \\ \ejc{I don't buy this argument. It's not because people use it that we should. People do lots of things and this does not mean that we should imitate them. I have some reservation about the value of predicting crime rate. What knowledge does this bring? What value does it bring? In which way is the world improved? I think we seriously need to think about all this and of the value of the things we {\bf choose} to do. (I am speaking here generally and not about this paper.) \\
% Let us return to the paper. Perhaps, we can use this data but we should explain why it might be useful to predict crime rates and why equalized odds makes sense in this context. (1) It might be useful to predict crime rate because society may want to think about resource allocation (police, economic relief and so on). This is not my area of expertise so I am making things up. (2) It actually may be a good thing to use equalized odds in this context for otherwise, we could potentially leave the door open to all sorts of biases? I think we need clarity on this.\\
% Please forgive my sensitivity and my harshness. They are not directed at you. I would just like to make sure we know why we do certain things. If we cannot provide a justification, then it means that we are doing arbitrary things.}

\smallskip

% \ejc{There was something that bothered me a bit about the paper and I could not really put my finger on it but I think I now can. The issue is that I feel we need clarity on one thing: what does it mean to obey EO (1)? (1) is too abstract. Do we mean that we obey a version of (1) on a future test set where $\hat f$ is fixed (fitted on an independent set)? Or do we mean that we want this to hold on the training set as Algorithm 1 seems to imply? I left some comments below and I feel we need clarity. }

\section{Introduction}

Machine learning algorithms are now frequently used to inform high-stakes decisions---and even to make them outright. As such, society has become increasingly critical of the ethical implications of automated decision making, and researchers in algorithmic fairness are responding with new tools. {While fairness is context dependent and may mean different things to different people}, a suite of recent work has given rise to a useful vocabulary for discussing fairness in automated systems \cite{dwork2012fairness, chouldechova2017fair, kleinberg2017inherent, barocas2017fairness, chouldechova2018frontiers}. Fairness constraints can often be articulated as conditional independence relations, and in this work we will focus on the \emph{equalized odds} criterion \citep{hardt2016equality}, defined as 
\begin{equation} \label{eq:eqodds_def}
\hat{Y} \independent A \mid Y
\end{equation}
{where the relationship above applies to test points}; here, $Y$ is the response variable, $A$ is a sensitive attribute (e.g. gender), $X$ is a vector of features that may also contain $A$, and  $\hat{Y} = \hat{f}(X)$ is the {prediction} obtained with a {{\em fixed}} prediction rule $\hat{f}(\cdot)$. While the idea that a {prediction rule obeying} the equalized odds property is desirable has gained traction, actually finding such {a rule} for a real-valued or multi-class response is a relatively open problem. Indeed, there are only a few recent works attempting this task \cite{zhang2018mitigating,mary2019fairness}. Moreover, there are no existing methods to rigorously check whether {a learned model achieves this property.}

% Machine learning algorithms are now frequently used to inform or make outright high-stakes decisions. As such, society had become increasingly critical of the ethical implications of automated decision making, and researchers in algorithmic fairness are responding. While fairness is not well defined, a suite of recent work has given rise to a useful vocabulary for discussing fairness in automated systems \cite{dwork2012fairness, chouldechova2017fair, kleinberg2017inherent, barocas2017fairness, chouldechova2018frontiers}. Nonetheless, important debates remain unresolved and methodological gaps remain open. In particular, while the idea that a predictor satisfying the {\em equalized odds} property \citep{hardt2016equality} is in some sense fair has gained traction in the field, actually finding such predictors is relatively open problem, with only a few recent works attempting this task. Moreover, there are no existing techniques to rigorously checking whether a predictor satisfies this property.

% Nonetheless, important debates remain unresolved and methodological gaps remain open. In particular, while the idea that a predictor satisfying the {\em equalized odds} property \citep{hardt2016equality} is in some sense fair has gained traction in the field, actually finding such predictors is relatively open problem, with only a few recent works attempting this task. Moreover, there are no existing techniques to rigorously checking whether a predictor satisfies this property.

We address these two questions by introducing a novel training scheme to fit models that approximately satisfy the equalized odds criterion and a hypothesis test to detect when a prediction rule violates this same criterion. Both solutions build off of one key idea: we create a synthetic version $\tilde A$ of the {sensitive} attribute such that the triple $(\hat Y, \tilde A, Y)$ obeys \eqref{eq:eqodds_def} with $\tilde A$ in lieu of $A$.  To achieve equitable model fits, we regularize our models toward the distribution of the synthetic data. Similarly, to test whether equalized odds holds, we compare the observed data to a collection of artificial data sets.
The synthetic data is straightforward to sample, making our framework both simple to implement and modular in that it {works together with any loss function, architecture, training algorithm, and so on. Based on real data experiments on both regression and multi-class classification tasks, we find improved performance compared to state-of-the-art methods.}

% Items to cover:
% \begin{itemize}
%         \item There are infinite many models that we can fit on data and make useful predictions. How to decide which model is desirable?
%         \item Some people believe it is desirable to force parity of some statistical measure across groups. As a case study, we focus on a popular measure in the literature on algorithmic fairness: equalized odds.
%         \item Challenge: not clear how to achieve such a fairness measures.
%         \item First contribution: we propose a flexible and effective framework to achieve equalized odds. Our methodology drives the fitted model to satisfy a desired form of conditional independence. We report that in some cases satisfying this measure does not affect much the prediction performance.
%         \item Second contribution: we offer statistical tests for conditional independence, used to verify that a fitted model indeed satisfies the required measure of fairness.
%         \item Caveat: even if the fitted model satisfies a desired measure of fairness, reporting a point estimate of the response does not communicate with decision makers the limits of predictive performance. How can we communicate reliably and effectively data analysis results?
%         \item Third contribution: the approach we take follows the method of equalized coverage. We offer a rigorous and operational framework to construct calibrated prediction sets to communicate the uncertainty in the prediction.
%         \item Remark: we only use data sets that are not binary, so there are fewer data sets to use.
% \end{itemize}{}

\subsection{A synthetic example} \label{sec:synth}

\begin{figure}[t]
\centering 
	\begin{subfigure}[a]{0.24\textwidth}
	\includegraphics[width=1\textwidth, trim={0 0.3cm 0 0},clip]{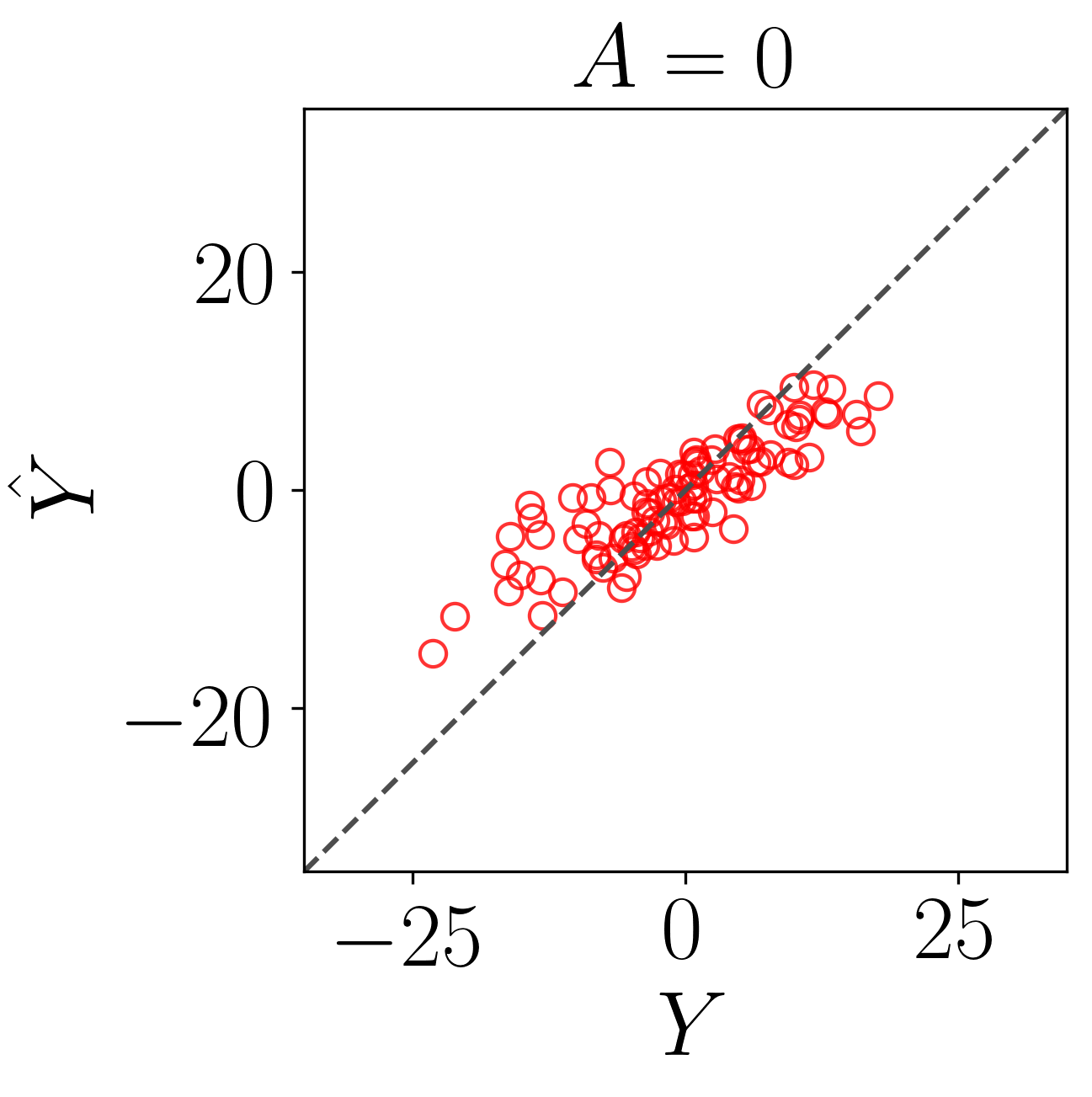}
	\caption{\small{\mbox{Baseline, minority.}}}
	\label{subfig:synth_baseline_0}
	\end{subfigure}
% 	\vspace{15pt}
% 	\hspace{1cm}
	\begin{subfigure}[a]{0.24\textwidth}
	\includegraphics[width=1\textwidth, trim={0 0.3cm 0 0},clip]{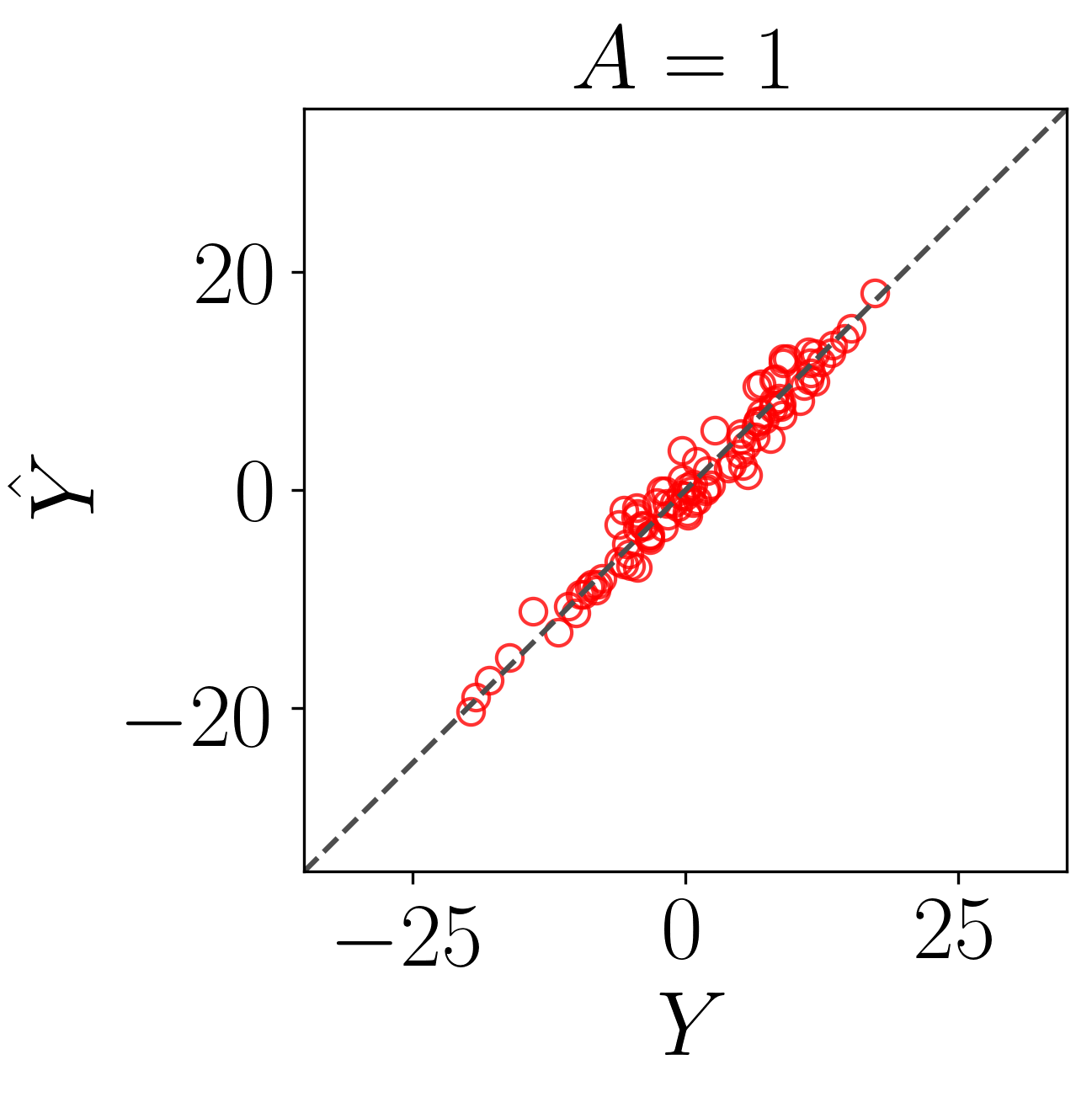}
	\caption{\small{\mbox{Baseline, majority.}}}
	\label{subfig:synth_baseline_1}
	\end{subfigure}
	\begin{subfigure}[a]{0.24\textwidth}
	\includegraphics[width=1\textwidth, trim={0 0.3cm 0 0},clip]{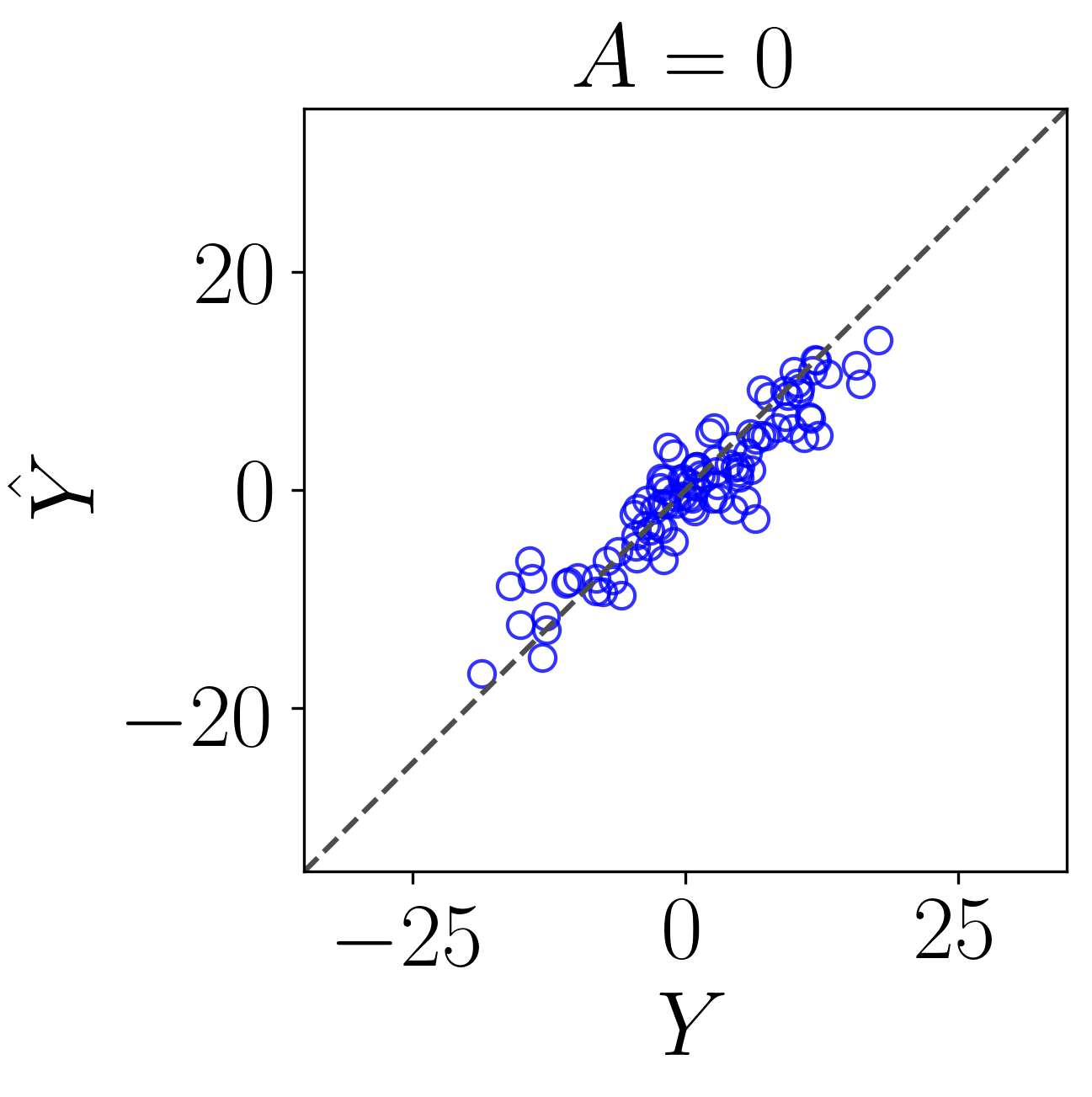}
	\caption{\small{\mbox{Proposed, minority.}}}
	\label{subfig:synth_equi_0}
	\end{subfigure}
% 	\vspace{15pt}
% 	\hspace{1cm}
	\begin{subfigure}[a]{0.24\textwidth}
	\includegraphics[width=1\textwidth, trim={0 0.3cm 0 0},clip]{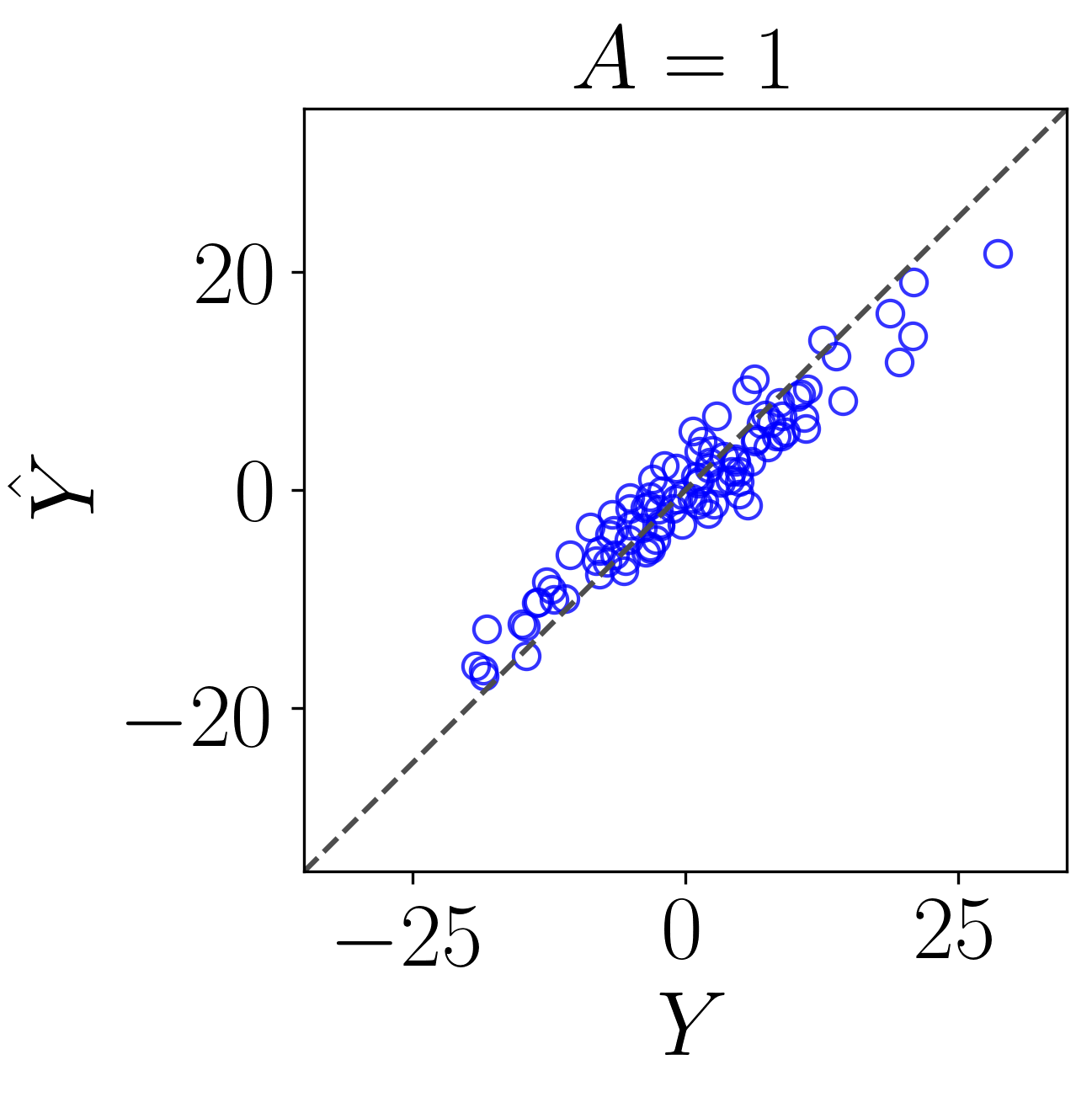}
	\caption{\small{\mbox{Proposed, majority.}}}
	\label{subfig:synth_equi_1}
	\end{subfigure}
	\caption{The effect of our learning framework on simulated data: (a,b) predictions from the baseline linear model; (c,d) predictions from the linear model fitted with the proposed equalized odds penalty. 
	%Red points in (a,b) and blue points in (c,d) represent the predictions $\hat{Y}$ obtained by the baseline and the equitable model, respectively.
	}
	\label{fig:synthetic_illustration}
\end{figure}

To set the stage for our methodology, we first present an experiment demonstrating the challenges of making equitable predictions as well as a preview of our method's results. We simulate a regression data set with a binary {sensitive}
%\ejc{I would use sensitive throughout}\yr{fixed}
attribute and two features:
\begin{equation}
    (X_1,X_2) \mid (A=0) \ \eqd \ (Z_1, 3Z_2) \quad \text{and} \quad  (X_1,X_2) \mid (A=1) \ \eqd \ (3Z_1, Z_2),
\end{equation}
where $Z_1,Z_2 \sim \mathcal{N}(0,1)$ is a pair of independent standard normal variables, and the symbol $\eqd$ denotes equality in distribution. {We create a population where 90\% of the observations are from the group $A=1$ in order to investigate a setting with a large majority group.} 
After conditioning on $A$, the model for $Y \mid X$ is linear: $Y~=~X^{\top}\beta_{A}~+~\epsilon,$
with noise $\epsilon \sim \mathcal{N}(0,1)$ and coefficients
$\beta_{0} = (0,3)$ and $\beta_{1} = (3,0)$. {We designed the model in this way so that the distribution of $Y$ given $X$ is the same for the two groups, up to a permutation of the coordinates. (In some settings, we might say that both groups are therefore equally deserving.) 
Consequently, the best model has equal performance in both groups. We therefore find it reasonable to search for a fitted model that achieves equalized odds in this setting.} 

To serve as an initial point of comparison, we first fit a classic linear regression model with coefficients $\hat{\beta}\in\RR^2$ on the training data, minimizing the mean squared error. Figures~\eqref{subfig:synth_baseline_0} and \eqref{subfig:synth_baseline_1} show the performance of the fitted model for each group {on a separate test set}. The fitted model performs significantly better on the samples from the majority group $A=1$ than those from the minority group $A=0$. {This is not surprising since the model seeks to minimize the overall prediction error.} Here, the overall {root} mean squared error {(RMSE)} {evaluated on test points} is equal to {2.29}, with an average value of {4.96} for group $A=0$ and of {1.79} for group $A=1$. {It is visually clear that for any vertical slice of the graph at a fixed value of $Y$, the distribution of $\hat{Y}$ is different in the two classes, i.e. the equalized odds property in \eqref{eq:eqodds_def} is violated.} 
%\ejc{Say a bit more here, e.g. that the distribution of $\hat Y \mid Y$ is very different in both groups}
This fact can be checked formally with our hypothesis test for \eqref{eq:eqodds_def} described later in Section \ref{sec:crt}. The resulting p-value {on the test set} is $0.001$ providing rigorous evidence that equalized odds is violated in this case.

Next, we apply our proposed fitting method (see Section~\ref{sec:achieving_eo}) on this data set. Rather than a naive least squares fit, we instead fit a linear regression model that approximately satisfies the equalized odds criterion. The new predictions are displayed in Figures~\eqref{subfig:synth_equi_0} and \eqref{subfig:synth_equi_1}. In contrast to the naive fit, the new predictive model achieves {a more balanced performance} across the two groups: the blue points are dispersed similarly in these two panels. This observation is consistent with the results of our hypothesis test; the p-value {on the test set} is equal to $0.452$, which provides no indication that the equalized odds property is violated. Turning to the statistical efficiency, the equitable model has improved performance for observations in the minority group $A=0$ with an {RMSE equal to 3.07}, at the price of reduced performance in the majority group $A=1$, where the {RMSE rises to 3.35}. The overall RMSE is 3.33, larger than that of the baseline model.
%\ejc{I would report the RMSE rather than the RMSE since this would put the numbers on the right unit scale.}
\subsection{Related work}\label{sec:related_work}

%\ejc{I will repeat myself. I do not like the word 'protected'. Unless you prove me wrong, I think we mostly discuss applications where there is no law affirming the role of protected attributes.} \yr{fixed}
%equalized odds, how to achieve
The notion of equalized odds as a criterion for algorithmic fairness was introduced in \cite{hardt2016equality}. In the special case of a binary target variables and a binary response variable, the aforementioned work offered a procedure to post-process any predictive model to construct a new model achieving equalized odds, possibly at the cost of reduced accuracy. Building off this notion of fairness, \cite{zafar2017fairnessconstraints} and \cite{donini2018empirical} show how to fit linear and kernel classifiers that are aligned with this criterion as well---these methods apply  when the response and sensitive attribute are both binary. Similarly, building on the Hirschfeld-Gebelein-Renyi (HGR) Maximum Correlation Coefficient, \cite{mary2019fairness} introduces a penalization scheme to fit neural networks that approximately obey equalized odds, applying to continuous targets and {sensitive} attributes. Coming at the problem from a different angle, \cite{louppe2017learning,zhang2018mitigating} fit  models with an equalized odds penalty using an adversarial learning scheme. The main idea behind this method is to maximize the prediction accuracy while minimizing the adversary’s ability to predict the {sensitive} attribute. {Our method has the same objective as the latter two, but  uses a new subsampling technique for regularization, which also leads to the first formal test of the equalized odds property in the literature.} 

%conformal inference
%Quantifying uncertainty while certifying fairness properties is less developed in the literature. \cite{romano2019malice} show how calibrate any any prediction rule to return predictive sets that have the same accuracy across strata of the protected attribute. Their uncertainty quantification leverages the ideas of conformal inference \citep{vovk1999machine, vovk2005algorithmic, romano2019conformalized}.

\section{Fitting fair models}
\label{sec:achieving_eo}

\subsection{Regularization with fair dummies} \label{subsection:regularization}

This section presents a method for fitting a predictive function $\hat{f}(\cdot)$ on i.i.d.~training data $\{(X_i, A_i, Y_i)\}$ indexed by $i \in \mathcal{I}_{\mathrm{train}}$ that approximately satisfies the equalized odds property \eqref{eq:eqodds_def}. {In regression settings, {let} $\hat{Y}=\hat{f}(X) \in \RR $ {be the predicted value of} the continuous response $Y \in \RR$. In multi-class classification problems where the response variable $Y \in \{1,\dots,L\}$ is discrete, we take the output of the classifier to be $ \hat{Y} = \hat{f}(X) \in \RR^L$, {a vector whose entries are estimated probabilities that an observation with $X=x$ belongs to class $Y=y$.}}
%\ejc{I am not sure I understand the dichotomy. Why do we provide an estimate of the conditional distribution in one case and only a prediction in the other? I find this very strange. We'd need to justify this.}
{We use this formulation of $\hat{Y}$ because it is the typical information available to the user when deploying a neural network for regression or classification, and our methods will use neural networks as the underlying predictive model. Nonetheless, the material in this subsection holds for any formulation of $\hat{Y}$, such as an estimated class label.}

%For convenience, we assume that the estimated class probabilities are normalized to sum to one.

Our procedure starts by constructing a \emph{fair dummy} {sensitive} attribute $\tilde{A}_i$ for each training sample:
\begin{align}
\tilde{A}_i \sim P_{A|Y}\left(A_i \mid Y_i\right), \quad i \in \mathcal{I}_{\mathrm{train}}, %i=1,2,\dots,n,
\end{align}
%\ejc{I don't get $y_i$.}\yr{$y_i$ is the true response of the $i$th examples (a scalar quantity)}
where $P_{A|Y}$ denotes the conditional distribution of $A_i$ given $Y_i$. This sampling is straightforward; see \eqref{eq:sampling_dummies} below.
Importantly, we generate $\tilde{A}_i$ without looking at $\hat{Y}_i$ so that we have the following property:
\begin{align}\label{eq:control}
 \hat{Y}_i \independent \tilde{A}_i \mid Y_i, \quad  i \in \mathcal{I}_{\mathrm{train}}. 
 %i=1,2,\dots,n.
\end{align} 
Notice that the above is exactly the equalized odds relation in \eqref{eq:eqodds_def}, with a crucial difference that the original {sensitive} attribute $A_i$ is replaced by the artificial one $\tilde{A}_i$. We will leverage this fair, synthetic data for both model fitting and hypothesis testing in the remainder of this work.

{Motivated by} \eqref{eq:control}, we propose the following objective function for equalized odds model fitting:
% \begin{equation} \label{eq:epi_optimization}
% \hat{f}(x) = \underset{f \in \mathcal{F}}{\mathrm{argmin}} \ \frac{1-\lambda}{n}\sum_{i=1}^n \ell(Y_i,f(X_i)) + \lambda \mathcal{D}\left( ( \hat{\Y}, \A, \Y ), ( \hat{\Y}, \tilde{\A}, \Y ) \right).
% \end{equation}
\begin{equation} \label{eq:epi_optimization}
\hat{f}(x) = \underset{f \in \mathcal{F}}{\mathrm{argmin}} \ \frac{1-\lambda}{|\mathcal{I}_{\mathrm{train}}|}\sum_{i \in \mathcal{I}_{\mathrm{train}} } \ell(Y_i,f(X_i)) + \lambda \mathcal{D}\left( ( \hat{\Y}, \A, \Y ), ( \hat{\Y}, \tilde{\A}, \Y ) \right).
\end{equation}
Here, $\ell(\cdot)$ is a loss function that measures the prediction error, such as the mean squared error for regression, or the cross-entropy for multi-class classification. The second term on the right hand side is a penalty promoting the equalized odds property, described in detail soon. The hyperparameter $\lambda$ trades off accuracy versus equalized odds. Above, the $i$th row of $\hat{\Y}\in\RR^{|\mathcal{I}_{\mathrm{train}}| \times k}$ is $f(X_i) \in \R^k$, {with $k=1$ in regression and $k=L$ in multi-class classification.} %\ejc{Why does $f(X) \in \mathbb{R}^k?$ I may not get this.}\yr{There is no way you could understand it. Now, at the beginning of this section we describe the regression and multi-class classification settings.} 
Similarly, we define $\mathbf{X}\in \RR^{|\mathcal{I}_{\mathrm{train}}| \times p}$ $\A\in\RR^{|\mathcal{I}_{\mathrm{train}}|}$, $\tilde{\A}\in\RR^{|\mathcal{I}_{\mathrm{train}}|}$, and $\Y\in\RR^{|\mathcal{I}_{\mathrm{train}}|}$,
whose entries correspond to the features, {sensitive} attributes, fair dummies, and labels, respectively.
% \begin{align}
%         \mathbf{A} = \begin{bmatrix}
%           A_{1},
%           A_{2},
%           \cdots,
%           A_{|\mathcal{I}_{\mathrm{train}}|}
%          \end{bmatrix}^\top, \ \ \ \tilde{\mathbf{A}} = \begin{bmatrix}
%           \tilde{A}_{1}, 
%           \tilde{A}_{2},
%           \dots,
%           \tilde{A}_{|\mathcal{I}_{\mathrm{train}}|}
%          \end{bmatrix}^\top, \ \ \ \mathbf{Y} = \begin{bmatrix}
%           Y_{1},
%           Y_{2},
%           \dots,
%           Y_{|\mathcal{I}_{\mathrm{train}}|}
%          \end{bmatrix}^\top,
% \end{align}
% are $n$-dimensional column vectors whose entries correspond to the {sensitive} attributes, fair dummies, and labels, respectively.
As a result, both $(\hat{\Y}, \A, \Y)$ and $(\hat{\Y}, \tilde{\A}, \Y)$ are matrices of size $|\mathcal{I}_{\mathrm{train}}| \times (k + 2)$. The function $\mathcal{D}(\U,\V)$ is any measure of the discrepancy between two probability distributions $P_{U}$ and $P_{V}$ based on the samples $\U$ and $\V$, summarizing the differences between the two samples into a real-valued score. A large value suggests that $P_U$ and $P_V$ are distinct, whereas a small value suggests that they are similar. We give a concrete choice based on adversarial classification in Section \ref{sec:impl_two_sample}. Since $(\hat{\Y}, \tilde{\A}, \Y )$ obeys the equalized odds property by construction, making the discrepancy with $(\hat{\Y}, \A, \Y )$ small forces the latter to approximately obey equalized odds.

% \ejc{I want to make changes here. I don't like the way the proposition is phrased and I also think it is incomplete. At the moment the prop says that (P) $(Y,A,\hat{Y}) \eqd (Y, \tilde A, Y)$ implies (EO). But the converse is true so that (EO) holds iff (P) holds. This is important since we introduce a p-value for (P). Without a converse, this may not be a p-value for (EO). Agreed? If so, please add a proof in the SM right away and provide a pointer in the mss.}

\begin{prop}
{Take $(X, A, Y) \sim P_{XAY}$ and set $\hat{Y} = \hat{f}(X)$ for some fixed $\hat{f}(\cdot)$ (again, $X$ may include $A$). Let $\tilde  A$ be sampled indpendently from $P_{A|Y}(A|Y)$.\footnote{This means that we can write $\tilde{A} = h(Y, \epsilon)$ for some function $h(\cdot)$, where the random variable $\epsilon$ is independent of everything else.} Then, $\hat Y \independent A \mid Y$ if and only if
$(Y,A,\hat{Y}) \eqd (Y, \tilde A, Y)$. 
}
% $i \in \mathcal{I}_{\m$ \ejc{I don't understand why this must be $\mathcal{I}_{\mathrm{train}}$. I think it should be a generic $\mathcal{I}$. Also we do not want to confuse whether $\hat f$ is fitted on the same set or not.} 
% are i.i.d. from $P_{XAY}$. \rev{Set $\hat{Y}_i = \hat{f}(X_i)$ for some {\em fixed} function $\hat{f}(\cdot)$ and resample $\tilde{A}_i~\sim~P_{A|Y}\left(A_i \mid Y_i\right)$ independently from each other.\footnote{This means that we can write $\tilde{A}_i = f(Y_i, \epsilon_i)$, where $\{\epsilon_i\}$ is an i.i.d.~sequence independent of everything else.}  If $(\hat{\Y}, \A, \Y) \eqd (\hat{\Y}, \tilde{\A}, \Y) $, then $\mathbf{\hat{Y}} \independent \mathbf{A} \mid \mathbf{Y}$.}
% Suppose further that $(\hat{\Y}, \A, \Y) \eqd (\hat{\Y}, \tilde{\A}, \Y) $ with $\tilde{A}_i~\sim~P_{A|Y}\left(A_i \mid Y_i\right)$ conditionally independent from {each other} and from $\mathbf{X}$ and $\mathbf{A}$ given $\mathbf{Y}$ and where $\hat{Y}_i = \hat{f}(X_i)$ for some fixed function $\hat{f}(\cdot)$.  Then 
% $\mathbf{\hat{Y}} \independent \mathbf{A} \mid \mathbf{Y}$.
\label{prop:fair_dummies_eo}
\end{prop}

The proof of this proposition as well as all other proofs are in 
Appendix~\ref{app:proofs}. We argue that this equivalence is particularly fruitful: indeed, if we find a prediction rule  $\hat{f}(\cdot)$ such that $(\hat{\Y}, \A, \Y )$ has the same distribution as $(\hat{\Y}, \tilde{\A}, \Y )$ (treating the prediction rule as fixed), then $\hat{f}(\cdot)$ exactly satisfies equalized odds. Motivated by this, our penalty drives the model to a point where these two distributions are close based on the training set.
%Of course, we do expect to achieve exact equality in practice when fitting a complex model with the above strategy.
When this happens, then, informally speaking, we expect that equalized odds approximately holds for future observations.

The regularization term in \eqref{eq:epi_optimization} can be used with essentially any existing machine learning framework, allowing us to fit a predictive model that is aligned with the equalized odds criterion, no matter whether the response is discrete, continuous, or multivariate. It remains to formulate an effective discriminator $\mathcal{D}(\cdot)$ to capture the difference between the two distributions, which we turn to next.

\subsection{The discrepancy measure}\label{sec:impl_two_sample}

A good discrepancy measure $\mathcal{D}(\cdot)$ should detect differences in distribution between the training data and the fair dummies in order to better promote equalized odds. Many examples have already been developed for the purpose of two-sample tests; examples include the Friedman-Rafsky test \cite{friedman1979multivariate}, the popular maximum mean discrepancy (MMD) \cite{gretton2012kernel}, the energy test \cite{szekely2013energy}, and classifier two-sample tests \cite{friedman1983graph,lopez2016revisiting}. The latter are tightly connected to the idea of generative adversarial networks \cite{goodfellow2014generative} which serves as the foundation of our procedure. 

To motivate our proposal, suppose we are given two independent data sets $\{U_i\}$ and $\{V_i\}$: the first contains samples of the form $U_i = (\hat{Y}_i, A_i, Y_i)$, and the second includes $ V_i = (\hat{Y}_i, \tilde{A}_i, Y_i)$. Our goal is to design a function that can distinguish between the two sets, so we assign a positive (resp.~negative) label to each $U_i$ (resp.~$V_i$) and fit a binary classifier $\hat{d}(\cdot)$. Under the null hypothesis that $P_U = P_V$, the classification accuracy of $\hat{d}(\cdot)$ on hold-out points should be close to $1/2$, while larger values provide evidence against the null. To turn this idea into a training scheme, we repeat the following two steps: first, we fit a classifier $\hat{d}(\cdot)$ whose goal is to recognize any difference in distribution between $U$ and $V$, and second, we fit a prediction function $\hat{f}(\cdot)$ that attempts to ``fool'' the classifier $\hat{d}(\cdot)$ while also minimizing the prediction error. In our experiment, the function $\hat{d}(\cdot)$ is formulated as a deep neural network with a differentiable loss function, so as the two models---$\hat{f}(\cdot)$ and $\hat{d}(\cdot)$---can be simultaneously trained via stochastic gradient descent.

While adversarial training is powerful, it can be sensitive to the choice of parameters and requires delicate tuning \cite{louppe2017learning,zhang2018mitigating}. To improve stability, we add an additional penalty that forces the relevant second moments of $U$ and $V$ to approximately match; {we penalize by $\|\cov(\hat{\mathbf{Y}}, \mathbf{A}) - \cov(\hat{\mathbf{Y}}, \tilde{\mathbf{A}})\|^2$ {where $\tilde{\mathbf{A}}$ is as in \eqref{eq:control} and} $\cov$ denotes the covariance, since under equalized odds this would be zero in the population {(because $(\hat Y, A) \eqd (\hat Y, \tilde{A})$ by Proposition \ref{prop:fair_dummies_eo})}.}
%For multi-class classification, we use a one-hot encoding for $Y$ in this previous expression so that it is invariant to relabeling of the classes.}
Combining all of the above elements, we can now give the full proposed procedure in Algorithm~\ref{alg:fit}.
%Combining all of the above elements, we can now give the full proposed procedure in Algorithm~\ref{alg:fit}. 
% \sdb{We will change this to partial correlation in the experiments and in the text if we have time}.

\begin{algorithm}
\textbf{Input}: Data $\{(X_i, A_i, Y_i)\}_{i\in\mathcal{I}_{\mathrm{train}}}$;
% \STATE Parameters: penalty weights $(\lambda, \gamma)$,  step size $\mu$, number of gradient steps $N_g$, and iterations $K$.
predictive model $\hat{f}_{\theta_f}(\cdot)$ and discriminator $\hat{d}_{\theta_d}(\cdot)$.

% \textbf{Procedure}
\begin{algorithmic}[1]
\FOR{$k = 1,\dots,K$}
\STATE Sample fair dummies $\tilde{A}_i \sim P_{A|Y}(A_i \mid Y_i), 
i \in \mathcal{I}_{\mathrm{train}}$.
%i=1,2,\dots,n$.
{See Section \ref{subsection:sampling} for details.}
\STATE Update the discriminator parameters $\theta_d$ by repeating the following for $N_g$ gradient steps:
\begin{align}
& \! \! \! \! \! \mathcal{J}_d(\theta_d) = \frac{1}{|\mathcal{I}_{\mathrm{train}}|}\sum_{i \in \mathcal{I}_{\mathrm{train}}} \bigg[ \log\left(\hat{d}_{\theta_d}\left(\hat{f}_{\theta_f}(X_i), A_i, Y_i\right)\right) + \log\left(1-\hat{d}_{\theta_d}\left(\hat{f}_{\theta_f}(X_i), \tilde{A}_i, Y_i\right)\right)\bigg] \\
& \! \! \! \! \! \theta_d  \leftarrow \theta_d - \mu \nabla_{\theta_d}\mathcal{J}_d(\theta_d)
\end{align}
\STATE Update the predictive model parameters $\theta_f$ by repeating the following for $N_g$ gradient steps:
\begin{align}
& \! \! \! \! \! \mathcal{J}_f(\theta_f) = \frac{1-\lambda}{| \mathcal{I}_{\mathrm{train}}|}\sum_{i \in \mathcal{I}_{\mathrm{train}}} \ell\left(Y
_i,\hat{f}_{\theta_f}(X_i)\right) + \lambda\gamma \|\cov(\hat{\mathbf{Y}}, \mathbf{A}) - \cov(\hat{\mathbf{Y}}, \tilde{\mathbf{A}})\|^2 \\
& \! \! \! \! \! \quad \quad \quad + \frac{\lambda}{| \mathcal{I}_{\mathrm{train}}|}\sum_{i \in \mathcal{I}_{\mathrm{train}}} \bigg[  \log\left(\hat{d}_{\theta_d}\left(\hat{f}_{\theta_f}(X_i), \tilde{A}_i, Y_i\right) \right) +
 \log\left(1-\hat{d}_{\theta_d}\left(\hat{f}_{\theta_f}(X_i), A_i, Y_i\right)\right)\bigg]  \\
& \! \! \! \! \! \theta_f  \leftarrow \theta_f - \mu \nabla_{\theta_f}\mathcal{J}_f(\theta_f)
\end{align}
%where $\Sigma$ and $\tilde{\Sigma}$ are the covariances of $\{(\hat{f}(X_i), A_i, Y_i)\}$ and $\{(\hat{f}(X_i), \tilde{A}_i, Y_i)\}$, respectively.
% \yr{For multi-class classification, we use a one-hot encoding of $Y \in \{1,\dots, L\}$ into $\{0, 1\}^L$, so the resulting dimensions of $\Sigma$ and $\tilde{\Sigma}$ are $(2L + 1) \times (2L + 1)$.}
% \ejc{Does this make sense? The covariance matrix seems completely dominated by the $2L \times 2L$ block $(Y, \hat Y)$.} \yr{Actually, it does not matter how we convert $Y$ and I am taking my words back. First, notice that the $2L \times 2L$ block $(Y, \hat Y)$ is the same for both covariances. Second, the terms $\langle \hat{A}_i,\tilde{A}_j \rangle$, $\langle {Y}_i,\tilde{A}_j \rangle$, and $\langle {Y}_i,{A}_j \rangle$ are constants from an optimization perspective. As a result, the factors that affect this discrepancy measure are $\langle \hat{Y}_i,\tilde{A}_j \rangle$ and $\langle \hat{Y}_i,{A}_j \rangle$} \sdb{We've added a clarification to this effect in section 2.2.}
\ENDFOR
\end{algorithmic}
\vspace{0.1cm}
\textbf{Output}: Predictive model $\hat{f}_{\theta_f}(\cdot)$ approximately satisfying equalized odds.
% \begin{algorithmic}
% \STATE 
% \end{algorithmic}
\caption{Fair Dummies Model Fitting}
\label{alg:fit}
\end{algorithm}

\subsection{Sampling fair dummies}\label{subsection:sampling}

To apply the proposed framework we must sample fair dummies $\tilde{A}$ from the distribution $P_{A|Y}$. Since this distribution is typically unknown, we use the training examples $\{(A_i,Y_i)\}_{i \in \mathcal{I}_{\mathrm{train}}}$ to estimate the conditional density of $A \mid Y$. For example, when the {sensitive} attribute of interest is binary, we apply Bayes' rule and obtain
\begin{align}
     \PP\{A=1|Y=y\} = \frac{\PP\{Y=y \mid A=1\}\PP\{A=1\}}{\PP\{Y=y \mid A=1\}\PP\{A=1\} + \PP\{Y=y \mid A=0\}\PP\{A=0\}}.
     \label{eq:sampling_dummies}
\end{align}
All the terms in the above equation are straightforward to estimate; in practice, we approximate terms of the form $\PP\{Y=y \mid A=a\}$ using a linear kernel density estimation.

\section{Validating equalized odds} \label{sec:crt}
% Explain the CRT
Once we have a fixed predictive model $\hat{f}(\cdot)$ in hand (for example, a model fit on a separate training set), it is important to carefully evaluate whether equalized odds is violated on test points $\{(X_i,A_i,Y_i)\}_{i\in \mathcal{I}_{\mathrm{test}}}$. To this end, we develop a hypothesis test for the relation \eqref{eq:eqodds_def}. Our test leverages once again the fair dummies $\Atilde_i$, but we emphasize that it applies to any prediction rule, not just those trained with our proposed fitting method. {The idea is straightforward: we generate many instances of the test fair dummies $\bAtilde$ and compare the observed test data $(\bYhat, \bA , \bY)$ to those with the dummy attributes $(\bYhat, \bAtilde , \bY)$, since the latter triple obeys equalized odds.} One can compare these distributions with any test statistic to obtain a valid hypothesis test; this is a special case of the conditional randomization test of \cite{candes2018panning}. In Algorithm~\ref{alg:crt} below, {we present a version of this general test using \cite{tansey2018holdout} to form test statistic based on a deep neural network $\hat{r}(\cdot)$.}
Invoking \cite{candes2018panning}, the output of the test is a p-value for the hypothesis that equalized odds holds:
%[Validity of the fair dummies test] 
\begin{prop}\label{thm:crt_valid}
{Suppose the test observations $(Y_i, X_i, A_i)$ for $i \in \mathcal{I}_{\mathrm{test}}$ are i.i.d.. {Set $\hat{Y}_i = \hat{f}(X_i)$ for a fixed function $\hat{f}(\cdot)$ and construct independently distributed fair dummies $\tilde{A}_i$ as in Proposition~\ref{prop:fair_dummies_eo}. If equalized odds holds for each $i$, i.e.,~$\hat Y_i \independent A_i \mid Y_i$, then the distribution of the output $p_v$ of Algorithm~\ref{alg:crt} stochastically dominates the uniform distribution; in other words, it is a valid p-value.}} 
\end{prop}
%\ejc{Perhaps provide a short proof in the SM.}

%CRT algorithm
\begin{algorithm}

\textbf{Input}: Data $\{(\hat{Y}_i, A_i, Y_i)\}$, $i \in \mathcal{I}_{\mathrm{test}}$

% \textbf{Procedure}
\begin{algorithmic}[1]
\STATE Split $\mathcal{I}_{\mathrm{test}}$ into disjoint subsets $\mathcal{I}_1$ and $\mathcal{I}_2$.
\STATE Fit a model $\hat{r}(A_i, Y_i)$ on $\{(\hat{Y}_i, A_i, Y_i): i\in \mathcal{I}_1\}$, aiming to predict $\hat{Y}_i$ given $(A_i, Y_i)$.
% Fit a test function $T(\cdot)$ on $\{(\hat{Y}_i, A_i, Y_i): i\in \mathcal{I}_1\}$.
% \STATE Compute the test statistic on the validation set: $t^* = T(\{(\hat{Y}_i, A_i, Y_i): i\in \mathcal{I}_2\})$. 
\STATE Compute the test statistic on the validation set: $t^* = \frac{1}{|\mathcal{I}_2|}\sum_{i\in\mathcal{I}_2} T(\hat{Y}_i,Y_i,\hat{r}(A_i, Y_i))$.
\FOR{$k = 1,\dots,K$}
\STATE Sample a fresh copy of the fair dummies $\tilde{A}_i \sim P_{A|Y}(A_i \mid Y_i), \ i\in \mathcal{I}_2$.
\STATE Compute the test statistic using the fair dummies: $t^{(k)} = \frac{1}{|\mathcal{I}_2|}\sum_{i\in\mathcal{I}_2} T(\hat{Y}_i,Y_i,\hat{r}(\tilde{A}_i, Y_i))$.
% $t^{(k)} = T(\{(\hat{Y}_i, \tilde{A}_i, Y_i): i\in \mathcal{I}_2\}).$ 
\ENDFOR
\STATE Compute the quantile of the true statistic $t^*$ among the fair dummy statistics $t_1, \dots, t_K$:
$$p_v = \frac{1 + \#\{k: t^* \le t^{(k)}\}}{ K+ 1}.$$
\end{algorithmic}

\textbf{Output}: A p-value $p_v$ for the hypothesis that \eqref{eq:eqodds_def} holds, valid under the assumptions of Proposition~\ref{thm:crt_valid}.
% \begin{algorithmic}
% \STATE 
% \end{algorithmic}
\caption{The Fair Dummies Test}
\label{alg:crt}
\end{algorithm}

%\subsection*{Implementation details}
%describe random forest test statistic (consider moving this to an appendix)

We reiterate that this holds for any choice of the test statistic $T(\cdot)$, so we next discuss a good all-around choice.
For problems with a continuous response $Y\in\RR$ {and prediction $\hat{Y}\in\RR$}, we define the test statistic as the squared error function, $
  T(\hat{Y}_i,Y_i,\hat{r}(A_i,Y_i)) = (\hat{Y}_i - \hat{r}(A_i,Y_i))^2.  
$
Here, $\hat{r}(\cdot)$ can be any model predicting $\hat{Y}_i \in \RR$ from $(A_i,Y_i)${; we use a two-layer neural network in our experiments.}
We describe a similar test statistic for multi-class classification in Appendix~\ref{app:multiclass_crt_stat}.
% ---we use a two-layer neural network in our experiments.\yr{``we use a two-layer...'' is redundant, I think this can be removed}

\section{Experiments} 
\label{sec:experiments}

We now evaluate our proposed fitting method in real data experiments. We compare our approach to two recently published methods, adversarial debiasing \cite{zhang2018mitigating} and HGR \cite{mary2019fairness}, demonstrating moderately improved performance.
%\footnote{Our implementation of adversarial debiasing is based on the software from \url{https://github.com/equialgo/fairness-in-ml}. We also adapt the software package of HGR to our setting, which is available online at \url{https://github.com/criteo-research/continuous-fairness}}
While our fitting algorithm also applies to binary classification, we only consider regression and multi-class classification tasks here because there are very few available techniques for such problems.
In all experiments, we randomly split the data into a training set (60\%), a hold-out set (20\%) to fit the test statistic for the fair-dummies test, and a test set (20\%) to evaluate their performance.
For reproducibility, all software is available at \url{https://github.com/yromano/fair_dummies}.

%Turning to a few details of the predictive intervals, in Sections~\ref{sec:experiments-reg}~and~\ref{sec:experiments-class} we randomly split the data into a proper training set (60\%), a calibration set (20\%) and a test set (20\%). We standardize the features to have zero mean and unit variance using the proper training set, and evaluate the performance over 10 different splits of the data. The miscoverage rate is fixed in all experiments and equal to $\alpha=0.1$.

\subsection{Real data: regression}\label{sec:experiments-reg}

\begin{figure}
\centering 
	\begin{subfigure}[a]{1\textwidth}
	\includegraphics[width=0.48\textwidth, trim={0.6cm 0.6cm 0.0cm 0.2cm},clip]{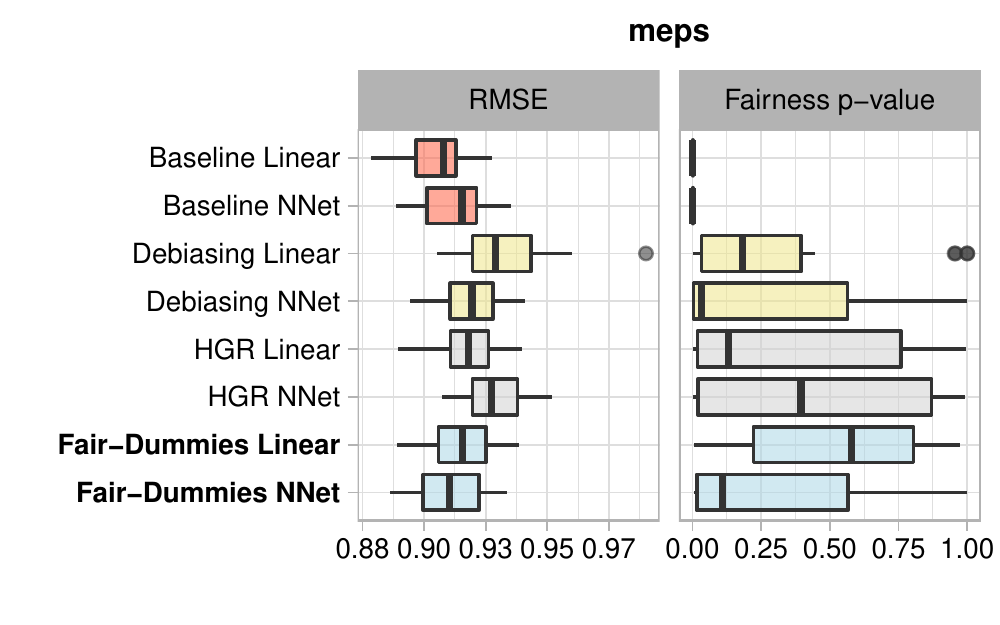}
	\includegraphics[width=0.48\textwidth, trim={0.6cm 0.6cm 0.0cm 0.2cm},clip]{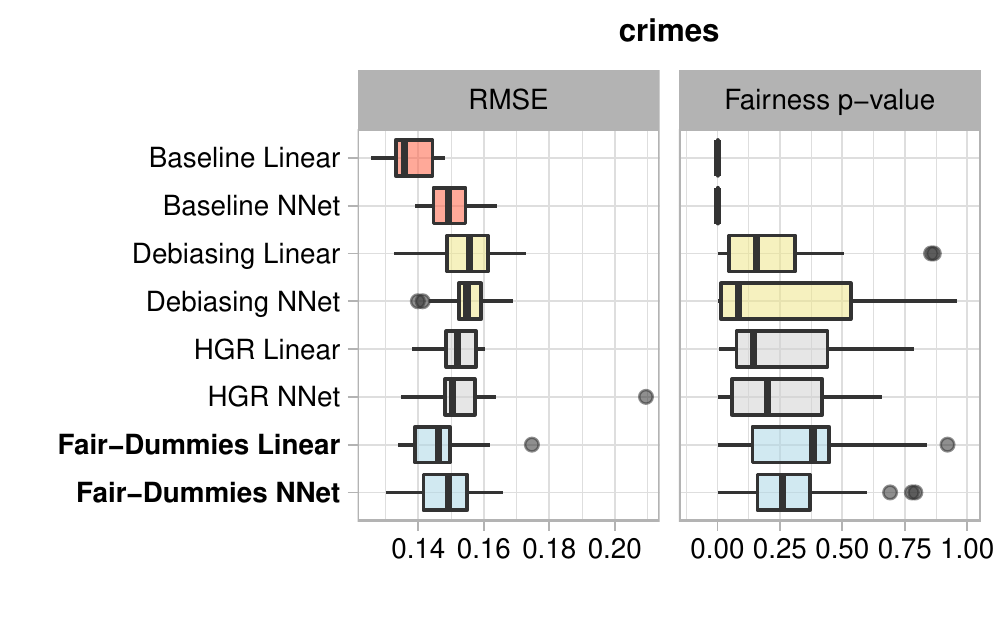}
	\label{subfig:crimes}
	\end{subfigure}
	\caption{Real data regression experiments on the MEPS (left) and Communities and Crimes (right) data sets. The results are shown over 20 random splits of the data. Each figure presents the {RMSE} as well as the equalized odds p-values obtained with the fair dummies test. 
	%Baseline algorithms (in red) are fit without any fairness penalty. In contrast, adversarial debiasing (yellow), HGR (gray), and our method (light blue) attempt to make predictions that satisfy the equalized odds criterion.
	}
	\label{fig:res_reg_mse}
\end{figure}

% \begin{figure}[t]
% \centering 
% 	\begin{subfigure}[a]{1\textwidth}
% 	\includegraphics[width=1\textwidth, trim={0.6cm 0.6cm 0.0cm 0.2cm},clip]{./meps_ec.pdf}
% 	\label{subfig:meps}
% 	\end{subfigure}
% % 	\hspace{1cm}
% 	\begin{subfigure}[a]{1\textwidth}
% 	\includegraphics[width=1\textwidth, trim={0.6cm 0.6cm 0.0cm 0.2cm},clip]{./crimes_ec.pdf}
% 	\label{subfig:crimes}
% 	\end{subfigure}
% 	\caption{Real data regression experiments on MEPS (top) and Communities and Crimes (bottom) data sets. The results are averaged over 10 random splits of the data. The target coverage level is $90\%$. Left to right: average prediction interval length per group, average coverage per group. Baseline algorithms (in red) are fit without any fairness penalty. In contrast, adversarial debiasing (yellow), HGR (gray), and our method (light blue) attempt to make predictions that satisfy the equalized odds criterion.}
% 	\label{fig:res_reg}
% \end{figure}

We begin with experiments on two data sets with real-valued responses: the 2016 Medical Expenditure Panel Survey (MEPS), where we seek to predict medical usage based on demographic variables, and the widely used UCI Communities and Crime data set, where we seek to predict violent crime levels from census and police data. See Appendix~\ref{app:data_regression} for more details. 
{Decision makers may wish to predict medical usage or crime rates to better allocate medical funding, social programs, police resources and so on \citep[e.g., ][]{henderson2010predicting}, but such information must be treated carefully.
 For both data sets we use race information as a binary {sensitive} attribute, and it is not used as a covariate for the predictive model. 
%  \ejc{This is good. To implement EO, we need to know race. Now do we use EO as a predictor variable? We need to say. If we do not use it, then ignore the rest. Otherwise, we need to say that we do not use race because we think it is causal. We merely use it to capture variability in the response that may not be captured by other variables. After this, we could use the remainder of your text. }\sdb{resolved}
An equalized odds model in this context can add a layer of protection against possible misuse of the model predictions by downstream agents: any two people (neighborhoods) with the same underlying medical usage (crime rate) would be treated the same by the model, regardless of racial makeup. Further care is still required to ensure that such a model is deployed ethically, but equalized odds serves as a useful safeguard.}

% \begin{itemize}
%     \item {CP Linear}: we fit a linear model by minimizing the mean squared error loss, and constructed the prediction interval using classic conformal prediction, described in Section~\ref{sec:classic_conformal}.
%     \item {CQR Linear}: a linear quantile regression model, whose output is a two-dimensional vector, representing the lower and upper conditional quantiles. The model is fitted by minimizing the pinball loss \eqref{eq:tot_pinball}, and conformalized by following the algorithm presented in Section~\ref{sec:cqr}.
%     \item {CP NNet}: a two layer neural network with a 64-dimensional hidden layer. The network is optimized by minimizing the mean squared error loss.
%     \item {CQR NNet}: the network architecture is the same as above, except that the output of the network is a two-dimensional vector, representing the two conditional quantiles. 
% \end{itemize}

We will consider two base predictors: a linear model and a neural network. As fairness-unaware baselines, we fit each of the above by minimizing the MSE, without any fairness promoting penalty. We also use each of the base regression models together with the {\em adversarial debiasing} method \cite{zhang2018mitigating}, the {\em HGR} method \cite{mary2019fairness}, and our proposed method;
see Appendix~\ref{app:learning} for technical details.
% To compare to existing methods that promote fairness, we consider adversarial debiasing \cite{louppe2017learning,zhang2018mitigating}: a recent method that builds upon adversarial training to mitigating violations of the equalized odds criterion. The core idea behind this method is to maximize the regressor's accuracy while minimizing the adversary’s ability to predict the \changed{sensitive} attribute.\footnote{Our implementation of adversarial debiasing relies heavily on the code from \url{https://github.com/equialgo/fairness-in-ml}} We also compare our method to that of \cite{mary2019fairness}, which uses a penalty which building on the Hirschfeld-Gebelein-Renyi (HGR) Maximum Correlation Coefficient to find a model approximately satisfying equalized odds.\footnote{Our implementation is based on the software package of HGR, available online at \url{https://github.com/criteo-research/continuous-fairness}}
The methods that promote equalized odds, including our own, each have many hyperparameters, and we find it challenging to automate the task of finding a set of parameters that maximizes accuracy while approximately achieving equalized odds, as also observed in \cite{zhang2018mitigating}.
%To quote the authors of the adversarial debiasing work \cite{zhang2018mitigating}: ``the adversarial training method is hard to get right and often touchy, in that getting the hyperparameters wrong results in quick divergence of the algorithm.'' 
Therefore, we choose to tune the set of parameters of each method only once and treat the chosen set as fixed in future experiments; see Appendix~\ref{app:hyper} for a full description of the tuning of each method. 
%To evaluate whether these methods satisfy the equalized odds requirement we apply the fair dummies test, presented in Algorithm~\ref{alg:crt}.

The performance of these methods is summarized in Figure~\ref{fig:res_reg_mse}. 
%We compare the MSE and the p-values from the fair dummies labeled as follows: Baseline (fairness-unaware model fitting), Debiasing~\cite{zhang2018mitigating}, HGR~\cite{mary2019fairness}, and Equitable (our proposal). 
We observe that the p-values of the two fairness-unaware baseline algorithms are small, indicating that the underlying predictions may not satisfy the equalized odds requirement. In contrast, adversarial debiasing, HGR, and our approach are all better aligned with the equalized odds criterion as the p-values of the fair dummies test are dispersed on the $[0,1]$ range. Turning to the predictive accuracy, we find that that the fairness-aware methods perform similarly to each other, although our proposed  
methods perform a little better
%model fitting with a linear predictor has slightly better performance
than the alternatives. Each of the fairness-aware models have slightly worse {RMSE} than the corresponding fairness-unaware baselines, as expected.

\subsection{Real data: multi-class classification}\label{sec:experiments-class}

{Next, we consider a multi-class classification example using the UCI Nursery data set, where we aim to rank nursery school applications based on family information. The response has four classes and we use financial standing as a binary {sensitive} attribute. See Appendix~\ref{app:data_classification} for more details.} Similar to our regression experiments, we use a linear multi-class logistic regression and neural network as fairness-unaware baseline algorithms. As before, we also fit predictive models using our proposed method and compare the results to those from adversarial debiasing and HGR. The latter only  handles one-dimensional $\hat{Y}$, so we adapted it to the multi-class setting by evaluating the penalty separately on each element of the vector of class-probabilities $\hat{Y}\in\RR^L$ and summing all $L$ of the penalty scores. See Appendix~\ref{app:learning} for additional details. 
%We implemented the fair dummies test by fitting $\hat{r}(\cdot)$ on the calibration set to formulate the test statistics \eqref{eq:crt_class}, see Section \ref{sec:crt} and Appendix~\ref{app:learning_details} for more details.  

%\begin{figure}[t]
\begin{wrapfigure}[16]{r}{0.5\textwidth}
\centering 
\vspace{-0.7cm}
	\includegraphics[width=0.49\textwidth, trim={0.6cm 0.6cm 0.0cm 0.2cm},clip]{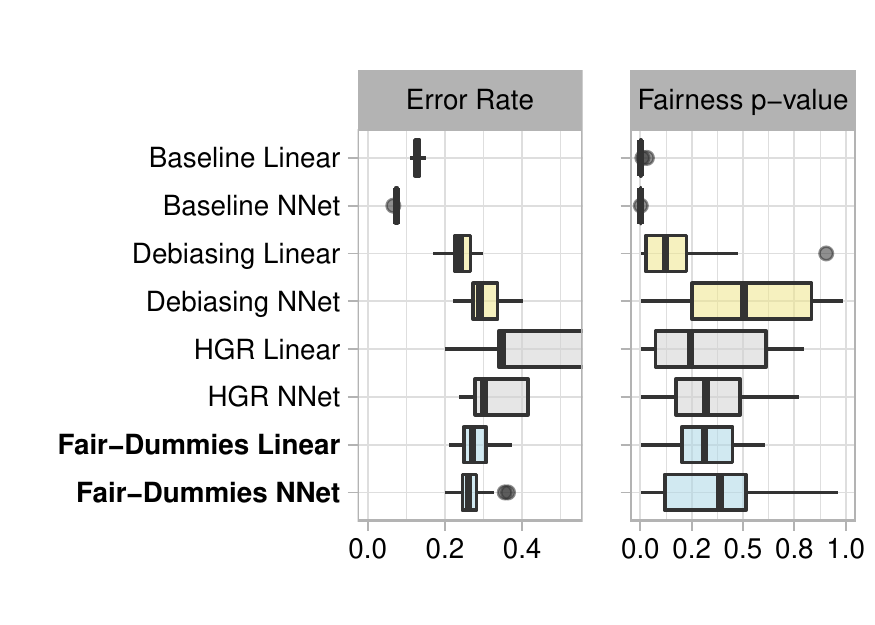}
	\label{subfig:nursery_test}
	\caption{Classification experiment on the Nursery data set. Results are shown for 20 random splits. Left: misclassification error. Right: fair dummies test p-value. Large values for HGR not shown.}
	\label{fig:res_class_acc}
\end{wrapfigure}
%\end{figure}

% \ejc{Please be careful about the positioning of the figures.}\yr{After removing the comments, we will make sure that all figures are aligned and that we are satisfying the page limit.}
We report the results in Figure~\ref{fig:res_class_acc}.
The p-values that correspond to the fairness-unaware baseline algorithms are close to zero, indicating that these methods violate the equalized odds requirement. In contrast, HGR, {adversarial debiasing, and our method} lead to a nice spread of the p-values over the $[0,1]$ range, with the exception of {adversarial debiasing with the linear model} which appears to violate equalized odds. Turning to the prediction error, when forcing the equalized odds criterion the statistical efficiency is significantly reduced compared to the fairness-unaware baselines, and since the linear {adversarial debiasing} method violates the equalized odds property,
%\ejc{which method is this?}
our method has the best performance among procedures that seem to satisfy equalized odds.

\section{Evaluating performance with uncertainty sets}
\label{subsec:eq_cov_sim}

Quantifying uncertainty in predictive modeling is essential, and, as a final case study, we revisit the previous data set with a new metric based on prediction sets. In particular, using the {\em equalized coverage} method \citep{romano2019malice}, we create predictive sets $C(X, A)\subseteq\{1,2,\dots,L\}$ that are guaranteed to contain the unknown response $Y$ with probability $90\%$. To ensure the prediction sets are unbiased to the {sensitive} attribute, the coverage property is made to hold identically across values of $A=a$:
\begin{equation}
\PP\{Y \in C(X,A) \mid A = a\} \geq 90\% \qquad \mathrm{for \ all } \ a \in \{0, 1\}.
\label{eq:eq_cov_intervals}
\end{equation}
Such sets can be created using any base predictor, and we report on these sets for the methods previously discussed in Figure~\ref{fig:res_class}; see Appendix~\ref{app:ec_details}.  We observe that all methods obtain exactly $90\%$ coverage per group, as guaranteed by the theory \cite{romano2019malice}. To compare the statistical efficiency, we look at the size of the prediction sets; smaller size corresponds to more precise predictions. 
%As in our previous experiments, requiring the equalized odds criterion degrades the statistical efficiency, resulting in larger prediction sets. 
Among the prediction rules that approximately satisfy equalized odds, a neural network trained with our proposed penalty performs the best (recall from Figure~\ref{fig:res_class_acc} that the linear method with adversarial debiasing violates equalized odds in this case).

\begin{figure}[t]
\centering 
	\begin{subfigure}[a]{1\textwidth}
	\includegraphics[width=1\textwidth, trim={0.6cm 0.6cm 0.0cm 0.2cm},clip]{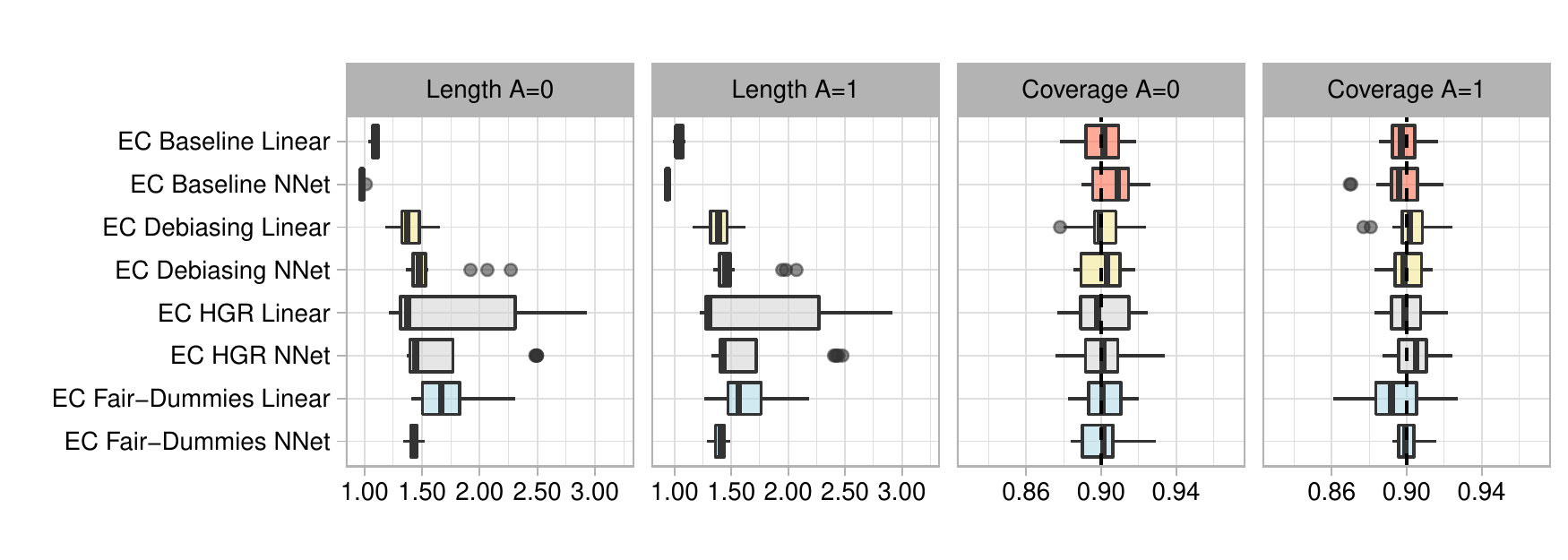}
	%\label{subfig:nursery_test}
	\end{subfigure}
	\caption{Classification experiment on the Nursery data set. The results are shown for 20 random splits. Left to right: average size of prediction set per group, coverage per group (target 90\%).}
	\label{fig:res_class}
\end{figure}

\section{Discussion}

In this work we presented a novel method for fitting models that approximately satisfy the equalized odds criterion, as well as a rigorous statistical test to detect violations of this property. The latter is the first of its kind, and we view it as an important step toward understanding the equalized odds property with complex models. Returning to the former, a handful of other approaches have been proposed, and we demonstrated similar or better performance to state-the-art methods in our numerical experiments. Beyond statistical efficiency, we wish to highlight the flexibility of our proposed approach. Our penalization scheme can be used with any discriminator or two sample test, any loss function, any architecture, any training algorithm, and so on, with minimal modification. Moreover, the inclusion of the second moment penalty makes our scheme stable, alleviating the sensitivity to the choice of hyperparameters.
{From a mathematical perspective, the synthetic data allows us to translate the problem of promoting and testing a conditional independence relation to the potentially more tractable problem of promoting and testing equality in distribution of two samples. {We expect this reframing will be useful broadly within algorithmic fairness. }}
%\ejc{I like this a lot!}
Lastly, we point out our procedure applies more generally to the task of fitting a predictive model while promoting a conditional independence relation \cite[e.g.,][]{louppe2017learning}, and leveraging this same technique in domains other than algorithmic fairness is a promising direction for future work.

We conclude with a critical discussion of the role of the equalized odds criterion in algorithmic fairness. We view our proposal as a way to {\em move beyond mean-squared error}; with modern flexible methods, there are often many prediction rules that achieve indistinguishable predictive performance, but they may have different properties with respect to robustness, fairness, and so on. When there is a rich enough set of good prediction rules, we can choose one that approximately satisfies the equalized odds property. Nonetheless, we point out two potential problems with exclusively focusing on the equalized odds criterion. First, it is well-known that forcing a learned model to satisfy the equalized odds can lead to decreased predictive performance \cite{kamiran2012data,feldman2015certifying,chouldechova2017fair,kleinberg2017inherent,menon2018cost, chen2018why}. Demanding that the equalized odds is exactly satisfied may force us to intentionally destroy information, as clearly seen in the algorithms for binary prediction rules in \cite{hardt2016equality,zafar2017fairnessconstraints,donini2018empirical,zhang2018mitigating,mary2019fairness}, and as implicitly happens in some of our experiments. Second, for regression and multi-class classification problems, there is no known way to certify that a prediction rule exactly satisfies equalized odds or to precisely bound the violation from this ideal, so the resulting prediction rules do not come with any formal guarantee. 
%It would be desirable to provide a rigorous finite-sample guarantee about a deployed prediction rule, but this seems to be out of reach when using complex machine learning algorithms to try to approximately satisfy the equalized odds criterion.
Both of these issues are alleviated when we return uncertainty intervals that satisfy the equalized coverage property, as shown in Section~\ref{subsec:eq_cov_sim}. With this approach, we regularize models towards equalized odds to the extent desired, while returning uncertainty sets valid for each group separately to accurately convey any difference in performance across the groups.
%and avoids intentionally destroying information about observations from the majority group.
Importantly, this gives an interpretable, finite-sample fairness guarantee only relying on the assumption of i.i.d. data. For these reasons, we see the combination of an (approximately) equalized odds model with equalized coverage predictive sets as an attractive combination for predictive models in high-stakes deployments.

\begin{ack}

E.~C. was partially supported by the Office of Naval Research grant N00014-20-12157, and by the National Science Foundation grants DMS 1712800 and OAC 1934578. He thanks Rina Barber and Chiara Sabatti for useful discussions related to this project. S.~B.~was supported by NSF under grant DMS 1712800 and a Ric Weiland Graduate Fellowship. Y.~R.~was supported by 
the Army Research Office (ARO) under
grant W911NF-17-1-0304. Y.~R.~thanks the Zuckerman
Institute, ISEF Foundation, the Viterbi Fellowship, Technion, and the Koret Foundation, for providing additional research support. 

\end{ack}

% \ejc{Some words in the refs are missing capitalization.} \yr{I could catch two cases, Stephen did I miss anything?} \sdb{I have not found any more. Emmanuel, let us know if there are more that you see.}

\bibliographystyle{unsrt}
\bibliography{MyBib}

\begin{thebibliography}{10}

\bibitem{dwork2012fairness}
Cynthia Dwork, Moritz Hardt, Toniann Pitassi, Omer Reingold, and Richard Zemel.
\newblock Fairness through awareness.
\newblock In {\em {Proceedings of the 3rd Innovations in Theoretical Computer
  Science Conference}}, pages 214--–226, 2012.

\bibitem{chouldechova2017fair}
Alexandra Chouldechova.
\newblock Fair prediction with disparate impact: A study of bias in recidivism
  prediction instruments.
\newblock {\em {Big Data}}, 5(2):153--163, 2017.

\bibitem{kleinberg2017inherent}
Jon Kleinberg, Sendhil Mullainathan, and Manish Raghavan.
\newblock Inherent trade-offs in the fair determination of risk scores.
\newblock In {\em {8th Innovations in Theoretical Computer Science Conference
  (ITCS 2017)}}, volume~67, pages 43:1--43:23, 2017.

\bibitem{barocas2017fairness}
Solon Barocas, Moritz Hardt, and Arvind Narayanan.
\newblock Fairness in machine learning.
\newblock {\em Advances in Neural Information Processing Systems Tutorial},
  2017.

\bibitem{chouldechova2018frontiers}
Alexandra Chouldechova and Aaron Roth.
\newblock A snapshot of the frontiers of fairness in machine learning.
\newblock {\em Commun. ACM}, 63(5):82–89, April 2020.

\bibitem{hardt2016equality}
Moritz Hardt, Eric Price, and Nati Srebro.
\newblock Equality of opportunity in supervised learning.
\newblock In {\em {Advances in Neural Information Processing Systems 29}},
  pages 3315--3323. 2016.

\bibitem{zhang2018mitigating}
Brian~Hu Zhang, Blake Lemoine, and Margaret Mitchell.
\newblock Mitigating unwanted biases with adversarial learning.
\newblock In {\em Proceedings of the 2018 AAAI/ACM Conference on AI, Ethics,
  and Society}, pages 335--340, 2018.

\bibitem{mary2019fairness}
J{\'e}r{\'e}mie Mary, Cl{\'e}ment Calauzenes, and Noureddine El~Karoui.
\newblock Fairness-aware learning for continuous attributes and treatments.
\newblock In {\em International Conference on Machine Learning}, pages
  4382--4391, 2019.

\bibitem{zafar2017fairnessconstraints}
Muhammad~Bilal Zafar, Isabel Valera, Manuel Gomez-Rodriguez, and Krishna~P.
  Gummadi.
\newblock Fairness constraints: A flexible approach for fair classification.
\newblock {\em {Journal of Machine Learning Research}}, 20(75):1--42, 2019.

\bibitem{donini2018empirical}
Michele Donini, Luca Oneto, Shai Ben-David, John~S Shawe-Taylor, and
  Massimiliano Pontil.
\newblock Empirical risk minimization under fairness constraints.
\newblock In {\em Advances in Neural Information Processing Systems 31}, pages
  2791--2801. 2018.

\bibitem{louppe2017learning}
Gilles Louppe, Michael Kagan, and Kyle Cranmer.
\newblock Learning to pivot with adversarial networks.
\newblock In {\em Advances in Neural Information Processing Systems 30}, pages
  981--990, 2017.

\bibitem{friedman1979multivariate}
Jerome~H Friedman and Lawrence~C Rafsky.
\newblock Multivariate generalizations of the {Wald-Wolfowitz} and {Smirnov}
  two-sample tests.
\newblock {\em The Annals of Statistics}, pages 697--717, 1979.

\bibitem{gretton2012kernel}
Arthur Gretton, Karsten~M Borgwardt, Malte~J Rasch, Bernhard Sch{\"o}lkopf, and
  Alexander Smola.
\newblock A kernel two-sample test.
\newblock {\em Journal of Machine Learning Research}, 13(Mar):723--773, 2012.

\bibitem{szekely2013energy}
G{\'a}bor~J Sz{\'e}kely and Maria~L Rizzo.
\newblock Energy statistics: A class of statistics based on distances.
\newblock {\em {Journal of Statistical Planning and Inference}},
  143(8):1249--1272, 2013.

\bibitem{friedman1983graph}
Jerome~H Friedman and Lawrence~C Rafsky.
\newblock Graph-theoretic measures of multivariate association and prediction.
\newblock {\em The Annals of Statistics}, 11(2):377--391, 1983.

\bibitem{lopez2016revisiting}
David Lopez{-}Paz and Maxime Oquab.
\newblock Revisiting classifier two-sample tests.
\newblock In {\em 5th International Conference on Learning Representations},
  2017.

\bibitem{goodfellow2014generative}
Ian Goodfellow, Jean Pouget-Abadie, Mehdi Mirza, Bing Xu, David Warde-Farley,
  Sherjil Ozair, Aaron Courville, and Yoshua Bengio.
\newblock Generative adversarial nets.
\newblock In {\em Advances in Neural Information Processing Systems 27}, pages
  2672--2680, 2014.

\bibitem{candes2018panning}
Emmanuel Cand\`es, Yingying Fan, Lucas Janson, and Jinchi Lv.
\newblock Panning for gold: Model-{X} knockoffs for high-dimensional controlled
  variable selection.
\newblock {\em Journal of the Royal Statistical Society: Series B},
  80(3):551--577, 2018.

\bibitem{tansey2018holdout}
Wesley Tansey, Victor Veitch, Haoran Zhang, Raul Rabadan, and David~M. Blei.
\newblock The holdout randomization test: Principled and easy black box feature
  selection, 2018.

\bibitem{henderson2010predicting}
Predicting crime.
\newblock {\em Arizona Law Review}, 52:15--64, 2010.

\bibitem{romano2019malice}
Yaniv Romano, Rina {Foygel Barber}, Chiara Sabatti, and Emmanuel Cand{\`e}s.
\newblock With malice toward none: Assessing uncertainty via equalized
  coverage.
\newblock {\em Harvard Data Science Review}, to appear.

\bibitem{kamiran2012data}
Faisal Kamiran and Toon Calders.
\newblock Data preprocessing techniques for classification without
  discrimination.
\newblock {\em Knowledge and Information Systems}, 33(1):1--33, 2012.

\bibitem{feldman2015certifying}
Michael Feldman, Sorelle~A. Friedler, John Moeller, Carlos Scheidegger, and
  Suresh Venkatasubramanian.
\newblock Certifying and removing disparate impact.
\newblock In {\em {Proceedings of the 21st ACM SIGKDD International Conference
  on Knowledge Discovery and Data Mining}}, pages 259--268, 2015.

\bibitem{menon2018cost}
Aditya~Krishna Menon and Robert~C Williamson.
\newblock The cost of fairness in binary classification.
\newblock In {\em Conference on Fairness, Accountability and Transparency},
  pages 107--118, 2018.

\bibitem{chen2018why}
Irene Chen, Fredrik~D Johansson, and David Sontag.
\newblock Why is my classifier discriminatory?
\newblock In {\em Advances in Neural Information Processing Systems 31}, pages
  3539--3550. 2018.

\bibitem{romano2019conformalized}
Yaniv Romano, Evan Patterson, and Emmanuel Cand{\`e}s.
\newblock Conformalized quantile regression.
\newblock In {\em {Advances in Neural Information Processing Systems 32}},
  pages 3543--3553. 2019.

\bibitem{kingma2014adam}
Diederik~P. Kingma and Jimmy Ba.
\newblock Adam: A method for stochastic optimization.
\newblock {\em arXiv preprint arXiv:1412.6980}, 2014.

\bibitem{vovk2005algorithmic}
Vladimir Vovk, Alex Gammerman, and Glenn Shafer.
\newblock {\em Algorithmic learning in a random world}.
\newblock Springer, 2005.

\bibitem{shafer2008tutorial}
Glenn Shafer and Vladimir Vovk.
\newblock A tutorial on conformal prediction.
\newblock {\em {Journal of Machine Learning Research}}, 9:371--421, 2008.

\end{thebibliography}

\appendix

\section{Proofs}
\label{app:proofs}

\begin{proof}[Proof of Proposition~1]
	The ``if'' direction is immediate. For the reverse direction, taking discrete random variables for simplicity, we have
	\begin{align*}
	\PP(\hat{Y} = \hat{y}, A = a, Y = y) &= \PP(\hat{Y} = \hat{y}, A = a \mid Y = y) \cdot \PP(Y = y) \\
	&= \PP(\hat{Y} = \hat{y} \mid Y = y) \cdot \PP(A = a \mid Y = y) \cdot \PP(Y = y) \\
	&= \PP(\hat{Y} = \hat{y} \mid Y = y) \cdot \PP(\tilde{A} = a \mid Y = y) \cdot \PP(Y = y) \\
	&= \PP(\hat{Y} = \hat{y}, \tilde{A} = a \mid Y = y) \cdot \PP(Y = y) \\
	&= \PP(\hat{Y} = \hat{y}, \tilde{A} = a, Y = y)
	\end{align*}
\end{proof}

\begin{proof}[Proof of Proposition~2]
	The proposed test is an instance of the Holdout Randomization Test \citep{tansey2018holdout}, which is in turn a special case of the Conditional Randomization Test \cite{candes2018panning}, so the result follows directly from Lemma 4.1 of \cite{candes2018panning}.
\end{proof}

\section{Test statistics for multi-class classification}
\label{app:multiclass_crt_stat}
{In this section, we give the details of the fair dummies test (Algorithm~\ref{alg:crt}) for multi-class classification.
	Here, with response $Y \in \{1,\dots,L\}$ and class probability estimates $\hat{Y} \in \RR^L$, let $\hat{Y}^{Y} \in \RR$ be the variable located in the $Y^{\text{th}}$ entry of $\hat{Y}$.}
% PREVIOUS VERSION
%For multi-class classification problems where the response variable $Y \in \{1,\dots,L\}$ is discrete, let $\hat{Y} = \hat{f}(X) \in \RR^L$ be a vector whose $y$'s entry is an estimate for the probability that an observation with $X=x$ belongs to class $Y=y$. For convenience, we assume that the estimated class probabilities are normalized to sum to one.
Similar to the regression case, we fit a predictive model {$\hat{r}({A}_i,Y_i) \in \RR$}, aiming to predict 
%the variable $\hat{Y}_i^{Y_i} \in \RR$ 
{the estimated class probability $\hat{Y}_i^{Y_i}$}
given the pair $(A_i,Y_i)$ by minimizing the cross entropy loss function. {(We use a one-hot encoding for $Y_i$.)} This function is then used to formulate our final test statistic: 
%\ejc{This is not clear to me. Is $\hat{Y}$ a vector? And what is $\hat Y^Y$? Is $\hat{r}$ a vector? This needs to be improved.}\yr{We defined the problems at the very beginning of Section \ref{subsection:regularization}. I believe it is clear now. In any case, $\hat{Y}\in\RR^L$ is a vector of class probabilities. $\hat{Y}^Y \in \RR$ is the $Y$'s entry in that vector. $\hat{r}(A,Y)\in\RR$ is a scalar, aiming to predict the class probability $\hat{Y}^Y$ } 
% $$T(\hat{Y}_i, Y_i, \hat{r}(A_i,Y_i)) = (\hat{Y}_i^{Y_i} - \hat{r}(A_i,Y_i))^2.$$
\begin{align}\label{eq:crt_class}
T(\hat{Y}_i, Y_i, \hat{r}(A_i,Y_i)) = -\hat{Y}_i^{Y_i}\log(\hat{r}(A_i,Y_i)) - (1-\hat{Y}_i^{Y_i})\log(1-\hat{r}(A_i,Y_i)).    
\end{align}
Another reasonable statistic for this setting would be to use the whole vector of class probabilities together with the multi-class cross-entropy loss, but we found that the above is more powerful at detecting violations of equalized odds. 
%\ejc{Is the test run on an independent data set? I.e. independent of that used to produce $\hat r$?}\yr{Yes. We stated in the main paper that we use the hold-out randomization test, as described in Algorithm \ref{alg:crt}}

\section{Data sets}

\subsection{Regression}\label{app:data_regression}

For regression problems, we compare the performance of our methods to adversarial debiasing \cite{zhang2018mitigating} and HGR \cite{mary2019fairness} on the following two data sets:
\begin{itemize}
	\item The 2016 Medical Expenditure Panel Survey (MEPS).\footnote{\url{https://meps.ahrq.gov/mepsweb/data_stats/download_data_files_detail.jsp?cboPufNumber=HC-192}} Here, the goal is to predict the utilization of medical services based on features such as the individual's age, marital status, race, poverty status, and functional limitations. {After pre-processing the data as in \cite{romano2019conformalized},} there are $15656$ samples and $138$ features. We take race as the binary {sensitive} attribute---there are $9640$ white individuals and $6016$ non-white individuals. {Note that MEPS data is subject to usage rules. We downloaded the data set using conformalized quantile regression \cite{romano2019conformalized} software package, available online.\footnote{\url{https://github.com/yromano/cqr}}}

	\item Communities and Crime data set.\footnote{\url{http://archive.ics.uci.edu/ml/datasets/communities+and+crime}} The goal is to estimate the number of violent crimes for U.S. cities given the median family income, per capita number of police officers, percent of officers assigned to drug units, and so on. {We clean the data according to \cite{mary2019fairness}, resulting in $1994$ observations of $121$ variables.} Race information is again used as the as {sensitive} attribute, with $784$ observations from communities whose percentage of African American is above 10\% and $1210$ observations from other communities.
\end{itemize}

\subsection{Multi-class classification}\label{app:data_classification}

The Nursery data contains information on nursery school applicants.\footnote{\url{https://archive.ics.uci.edu/ml/datasets/nursery}} The task is to rank applications based on features such as the parents' occupation, family structure, and financial standing. The original data set contains five classes, however, after cleaning and rearranging the data we remain with four classes: children who are (1) ``not recommended'', (2) ``very recommended'', (3) ``prioritized'', and (4) ``specifically prioritized'' to join the nursery. In total, the data set contains $12958$ examples and $13$ features. We use the financial status as a {sensitive} attribute; applicants with ``inconvenient'' standing are assigned to group $A=0$ ($6478$ samples) and those with ``convenient'' status are assigned to group $A=1$ ($6480$ samples).

\section{Further information about the learning algorithms} \label{app:learning}

\subsection{Hyper-parameter tuning}\label{app:hyper}
% \begin{figure}[t]
% \centering 
% 	\begin{subfigure}[a]{1\textwidth}
% 	\includegraphics[width=1\textwidth, trim={0.6cm 0.6cm 0.0cm 0.2cm},clip]{./figures/meps_train.pdf}
% 	\label{subfig:meps_train}
% 	\end{subfigure}
% % 	\hspace{1cm}
% 	\begin{subfigure}[a]{1\textwidth}
% 	\includegraphics[width=1\textwidth, trim={0.6cm 0.6cm 0.0cm 0.2cm},clip]{./figures/crimes_train.pdf}
% 	\label{subfig:crimes_train}
% 	\end{subfigure}
% 	\caption{Hyper-parameter tuning via 10 fold cross validation. Regression data sets: MEPS (top) and Communities and Crimes (bottom). Baseline: we choose the hyperparameters that minimize prediction error. Fairness methods (debiasing, HGR, and equitable): the hyperparameters are tuned to maximize statistical efficiency while passing the fair dummies test. The target coverage level per group is $90\%$.}
% 	\label{fig:res_reg_train}
% \end{figure}

% \begin{figure}[t]
% \centering 
% 	\begin{subfigure}[a]{1\textwidth}
% 	\includegraphics[width=1\textwidth, trim={0.6cm 0.6cm 0.0cm 0.2cm},clip]{./figures/nursery_train.pdf}
% 	\label{subfig:nursery_train}
% 	\end{subfigure}
% % 	\hspace{1cm}
% 	\caption{Hyper-parameter tuning via 10 fold cross validation for Nursery data set. See Figure \ref{fig:res_reg_train} for more details.}
% 	\label{fig:res_class_train}
% \end{figure}

To successfully deploy the learning algorithms presented in Section~\ref{sec:experiments}, we must tune various hyperparameters, such as the equalized odds penalty weight, learning rate, batch size, and number of epochs. This task is particularly challenging because we have a multi-criteria objective: the goal is not only to maximize accuracy but also to pass the fair dummies test, i.e. approximately achieve equalized odds. In our experiments, we find the best set of parameters using $10$ fold cross validation, optimizing the accuracy-fairness objective. 
%Figures \ref{fig:res_reg_train} and \ref{fig:res_class_train} summarize the resulting performance metrics, corresponding to the chosen set of hyperparameters per method, data set, and prediction task.  
Since this process is computationally expensive and partly manual, in practice, we tune the hyperparameters only once using cross validation on the entire data set and then treat the chosen set as fixed for the rest of the experiments. The drawback of this approach is that it may suffer from over-fitting, since we test on the same data used to tune the hyperparameters. To mitigate this problem, in Section \ref{sec:experiments}, we compare the performance metrics of the different algorithms on data splits that are different than the ones used to tune the parameters; some optimism, however, remains. In any case, this same tuning scheme is used for all methods, ensuring that the comparisons are meaningful.

\subsection{Implementation details} \label{app:learning_details}

\subsubsection*{Regression}

Our regression experiments build on two base learning algorithms, which are then combined with HGR, adversarial debiasing, and our framework to yield eight methods:
\begin{itemize}
	\item {Baseline Linear}: we fit a linear model by minimizing the MSE loss function, using the  stochastic gradient descent optimizer with a learning rate and number of epochs in $\{0.01, 0.1\}$ and $\{100, 200, 400, 600, 1000, 2000, 3000, 4000\}$, respectively. {We normalize the features to have zero mean and unit variance using the training data.}
	\item {Baseline NNet}: we fit a two layer neural network with a 64-dimensional hidden layer and ReLU nonlinearity function. The network is optimized by minimizing the MSE, following the same fitting strategy described in Baseline Linear. 
	\item Debiasing Linear and Debiasing NNet: the predictors are formulated as described in the baseline algorithms. Here, we follow the implementation provided in \url{https://github.com/equialgo/fairness-in-ml} and design the adversary as a four-layer neural network with hidden layers of size $32$ and ReLU nonlinearities. Since the { sensitive} attribute is binary, we apply the sigmoid function on the output of the last layer. We use the Adam optimizer \cite{kingma2014adam} for training, with a learning rate in $\{0.001, 0.01, 0.1\}$ and a minibatch size in $\{64, 128\}$. We also follow the pre-training strategy suggested in \cite{zhang2018mitigating} and fit separately the predictor and adversary for a number of epochs in $\{2, 4, 10, 20, 30, 40\}$. Then, the two pre-trained models are fitted interchangeably for additional $\{50, 100, 200, 300, 400\}$ epochs. The weight on the equalized odds penalty is selected from $\{0.2, 0.3, 0.4, 0.5, 0.6, 0.7, 0.8, 0.9, 0.95, 0.99\}$.
	\item HGR Linear and HGR NNet: we again use architectures identical to those of the baseline models. As suggested in \cite{mary2019fairness}, we use the Adam optimizer with a mini-batch size in $\{128, 256\}$, learning rate in $\{0.001, 0.01\}$, and the number of epochs in $\{10, 15, 20, 30, 40, 50, 80, 100\}$. The \texttt{HGR} function is implemented in \url{https://github.com/criteo-research/continuous-fairness} and we select the weight penalty from the $\{0.1, 0.2, 0.3, 0.4, 0.5, 0.6, 0.7, 0.8, 0.9\}$ range.
	\item Fair-Dummies Linear and Fair-Dummies NNet: we fit predictors that have the same structure as the baseline algorithms, with our proposed regularization. The discriminator is implemented as a two-layer neural network with a hidden layer of size $30$ and ReLU nonlinearities. We use the stochastic gradient descent optimizer, with a fixed learning rate of $0.01$. We use the same optimizer for the classifier, with the same learning rate, except for the addition of a momentum term with value $0.9$. The number of epochs is chosen from the $\{20, 30, 40 ,50, 80, 100\}$ range, and the number of gradient steps ($N_g$ in Algorithm~\ref{alg:fit}) is selected from the range of $\{40, 50, 60, 70, 80\}$. The weight on the equalized odds penalty is selected from $\{0.4, 0.5, 0.6, 0.7, 0.8, 0.9\}$ ($\lambda$ in Algorithm~\ref{alg:fit}), and the second moment term ($\gamma$ in Algorithm~\ref{alg:fit}) is chosen from $\{1, 10, 20\}$.
\end{itemize}
The predictive model $\hat{r}(\cdot)$, defining the test statistics in the fair dummies test (see Section \ref{sec:crt}), is formulated as a two-layer neural network, with a hidden dimension of size 64, and dropout layer with rate $1/2$. We use stochastic gradient descent to fit the network, run for $200$ epochs with a minibatch of size $128$ and a fixed momentum term with weight $0.9$. 

\subsubsection*{Multi-class classification}

Our experiments are again based on two underlying predictive models which are regularized using fairness-aware methodologies:
\begin{itemize}
	\item Baseline Linear: we fit a linear model by minimizing the cross entropy loss function. We use the Adam optimizer, with a minibatch size of 32. We choose the learning rate, and number of epochs from the range of $\{0.001, 0.01, 0.1\}$, and $\{20, 40, 60, 80, 100\}$, respectively. {We normalize the features to have zero mean and unit variance using the training data.}
	\item Baseline NNet: we fit a two layer neural network with a 64-dimensional hidden layer, ReLU nonlinearity function, and dropout regularization with rate $1/2$. We use the same optimization strategy as above above.
	\item Debiasing Linear and Debiasing NNet: we form classifiers as in the baseline algorithms. Similarly to the regression setting, we rely on the  implementation from \url{https://github.com/equialgo/fairness-in-ml}. We use the same adversary as described in the regression setting. Training is done via the Adam optimizer, with a fixed learning rate that is equal to $0.5$ and minibatches of size $32$. We again apply the pre-training strategy \cite{zhang2018mitigating} and fit separately the predictor and adversary for number of epochs from the range of $\{1, 2\}$. The adversarial training is then repeated for $\{20, 40, 60, 100, 200\}$ epochs. The weight on the equalized odds penalty is selected from $\{0.9, 0.99, 0.999, 0.9999, 0.99999, 0.999999\}$.
	\item HGR Linear and HGR NNet: we again take classifiers as in the baseline models. To fit them, we apply the Adam optimizer with a mini-batch size of $128$, learning rate in the range of  $\{0.001, 0.01\}$, number of epochs selected from $\{10, 20, 30, 40, 50\}$. The HGR penalty weight is selected in the range of $\{0.9, 0.91, 0.92, 0.93, 0.94, 0.95, 0.96, 0.97, 0.98, 0.99\}$.
	\item {Fair-Dummies Linear} and Fair-Dummies NNet: we again take classifiers with the same structure as the baseline algorithms. The adversary is implemented as a four-layer neural network with a 32-dimensional hidden layer and ReLU nonlinearity. We use the Adam optimizer, with a fixed learning rate that is equal to $0.5$. The number of epochs is fixed and equal to $50$. The number of gradient steps $N_g$ is selected in the range of $\{1,2\}$, and the weight on the equalized odds penalty is selected from $\{0.9, 0.99, 0.999, 0.9999, 0.99999, 0.999999\}$ for $\lambda$ and from $\{0.01, 0.001, 0001, 0.00001\}$ for $\gamma$.
\end{itemize}
The fair dummies test statistics is again evaluated using a predictive model $\hat{r}(\cdot)$ that is implemented as a neural network. We use the same architecture and learning strategy as in the regression setup, with the addition of a sigmoid function as the last layer. 

{
	\section{Further details on equalized coverage}\label{app:ec_details}
	
	We now turn to a few details of the equalized coverage prediction sets from Section~\ref{subsec:eq_cov_sim}. In our experiments, we use the software package provided by \cite{romano2019malice}, which is available online at \url{https://github.com/yromano/cqr}. While equalized coverage \cite{romano2019malice} is presented for regression problems, it is straightforward to extend this method to multi-class classification tasks. To this end, we follow split conformal prediction \cite{vovk2005algorithmic} and  randomly split the data into a proper training set (60\%), a hold-out calibration set (20\%), and a test set (20\%). We use the same predictive models from Section \ref{sec:experiments-class}, which are fitted to the whole proper training data, providing estimates for class probabilities. The examples $\{(X_i,A_i,Y_i)\}$ that belong to the calibration set are then used to construct the prediction sets for the test points. Specifically, following the notations from Section \ref{sec:crt}, we deploy the popular inverse probability conformity score \cite{shafer2008tutorial}, given by $1 - \hat{Y}_i^{Y_i}$. Here, $\hat{Y}_i=\hat{f}(X_i) \in \RR^L$ and the variable $\hat{Y_i}^{Y_i} \in \RR$ is the estimated probability that the calibration example $
	X_i$ belongs to class $Y_i$.

}

\end{document}